\documentclass{article}

\usepackage{microtype}
\usepackage{graphicx}

\usepackage{subcaption}
\usepackage{booktabs} 
\usepackage{xcolor} 
\usepackage{natbib}
\newcommand{\R}{\mathbb{R}}
\newcommand{\E}{\mathbb{E}}
\renewcommand{\P}{\mathbb{P}}
\newcommand{\calP}{{\cal P}}
\newcommand{\softmax}{\mbox{softmax}}
\usepackage{hyperref}

\usepackage{PRIMEarxiv}

\usepackage{amsmath}
\usepackage{amssymb}
\usepackage{mathtools}
\usepackage{amsthm}
\usepackage[inline]{enumitem}
\usepackage{dsfont}

\theoremstyle{plain}
\newtheorem{theorem}{Theorem}[section]
\newtheorem{proposition}[theorem]{Proposition}
\newtheorem{lemma}[theorem]{Lemma}
\newtheorem{corollary}[theorem]{Corollary}
\theoremstyle{definition}

\theoremstyle{remark}

\title{On The Statistical Representation Properties Of The Perturb-Softmax And The Perturb-Argmax Probability Distributions}

\author{%
  Hedda Cohen Indelman\\
  Technion\\
   \And
   Tamir Hazan \\
   Techion\\
}

\begin{document}

\maketitle

\begin{abstract}
The Gumbel-Softmax probability distribution allows learning discrete tokens in generative learning, while the Gumbel-Argmax probability distribution is useful in learning discrete structures in discriminative learning. Despite the efforts invested in optimizing these probability models, their statistical properties are under-explored. In this work, we investigate their representation properties and determine for which families of parameters these probability distributions are complete, i.e., can represent any probability distribution, and minimal, i.e., can represent a probability distribution uniquely. We rely on convexity and differentiability to determine these statistical conditions and extend this framework to general probability models, such as Gaussian-Softmax and Gaussian-Argmax. We experimentally validate the qualities of these extensions, which enjoy a faster convergence rate. We conclude the analysis by identifying two sets of parameters that satisfy these assumptions and thus admit a complete and minimal representation. Our contribution is theoretical with supporting practical evaluation.
\end{abstract}

\section{Introduction}
\label{sec:intro}

%\tamir{what is the problem and why is it important}

Learning over discrete probabilistic models is an active research field with numerous applications. Examples include learning probabilistic latent representations of semantic classes or beliefs \citep{10.5555/2969033.2969226, pmlr-v32-mnih14, pmlr-v5-salakhutdinov09a}. The Gumbel-Argmax and Gumbel-Softmax probability distributions are widely applied in machine learning to model and analyze discrete probability distributions. 

%\tamir{What people did and why is it not sufficient}

The Gumbel-Argmax is an equivalent representation of the softmax operation and plays a key role in ``follow the perturb-leader'' family of algorithms in online learning \citep{hannan1957approximation, kalai2002geometric, kalai2005efficient,NIPS2012_53adaf49}. Its extension to Gaussian-Argmax allows better bounds on their gradients and consequently provides better regret bounds in linear and high-dimensional settings \cite{ abernethy2014online, abernethy2016perturbation, pmlr-v37-cohena15}. The argmax operation allows for efficient sampling, making the Perturb-Argmax probability models pivotal in discriminative learning algorithms of high-dimensional discrete structures \citep{48965, Vlastelica2020Differentiation, Song2016TrainingDN, pmlr-v139-indelman21a, pmlr-v80-niculae18a}. The Gumbel-Softmax (or the Concrete distribution) probability distribution, which replaces the argmax operation with a softmax operation is easier to optimize and therefore plays a key role in generative learning models \citep{jang2017categorical, maddison2017}. The discrete nature of these probability models provides a natural representation of concepts, e.g., in 
zero-shot text-to-image generation (DALL-E) \citep{ramesh2021zero}. While the Gumbel-Argmax and Gumbel-Softmax probability distributions are widely applied in machine learning, their statistical representation is still under-explored. 

%\tamir{what we are doing and how we cure cancer}

\begin{figure*}[t]
  \centering
  \begin{subfigure}[t]{0.48\textwidth}
    \centering
    \includegraphics[width=.95\textwidth]{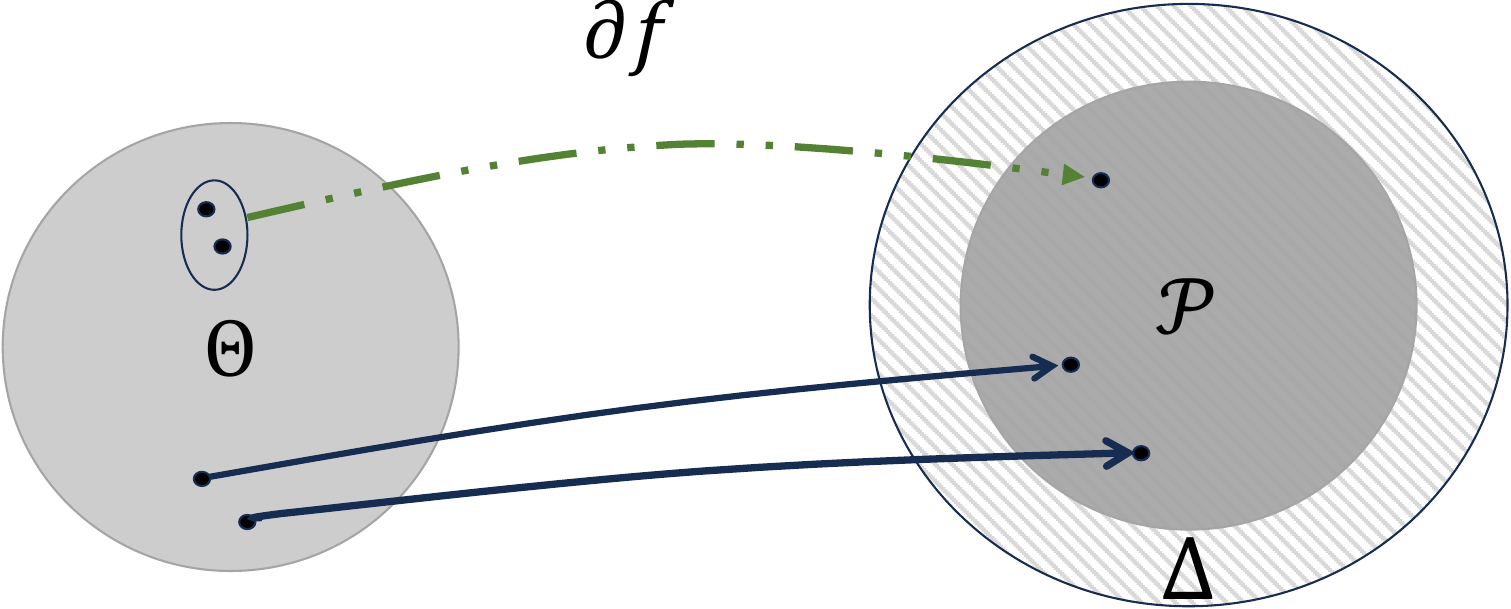}
    \caption{An illustration of the representation properties of the Perturb-Softmax. Its representation is complete for any perturbation distribution under the conditions in Theorem \ref{theorem:p_softmax_complete}. The mapping is one-to-one whenever the representation
    parameters are not linearly constrained (full mapping), and single-valued otherwise (green dashed mapping).}
\end{subfigure}%
 \hfill
  \begin{subfigure}[t]{0.48\textwidth}
    \centering
    \includegraphics[width=.95\textwidth]{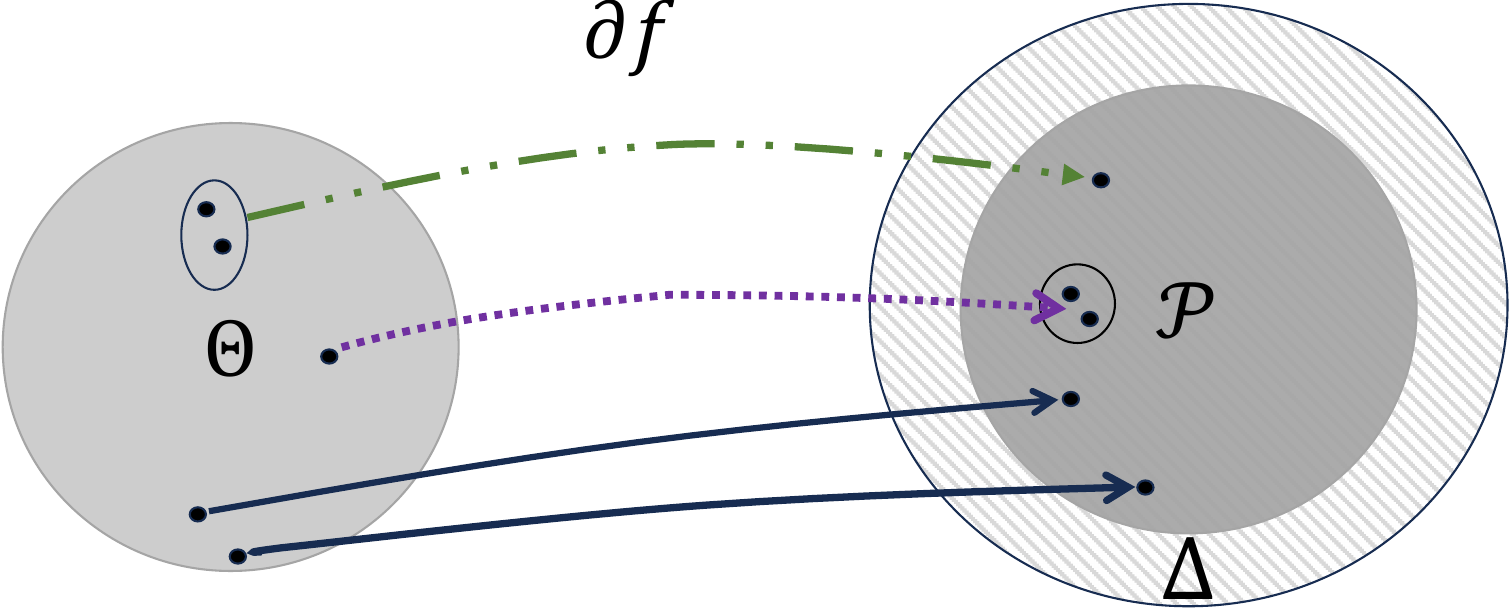}
    \caption{An illustration of the representation properties of the Perturb-Argmax. 
    Its representation is complete under the conditions in Theorem \ref{theorem:pert_max_complete}. The Perturb-Argmax is not identifiable whenever the perturbations follow a discrete distribution (dotted purple mapping), and identifiable for smooth pdf perturbation distributions (green dashed mapping). Moreover, under the conditions in Theorem \ref{theorem:p_argmax_minimal}, the mapping is one-to-one whenever the representation
    parameters are not linearly constrained (full mapping).}
\end{subfigure}
\caption{Illustration of the representation properties of the Perturb-Softmax and of the Perturb-Argmax.}
\label{fig:highlevel_properties}
\end{figure*}

In this work, we investigate the representation properties of the Gumbel-Argmax and Gumbel-Softmax probability distributions. We aim to determine for which families of parameters these distributions are complete and minimal. A distribution is considered complete if it can represent any probability distribution, and minimal if it can uniquely represent a probability distribution. Our statistical investigation realizes the Gumbel-Argmax and the Gumbel-Softmax probability distributions as gradients of respective convex functions. In Theorems \ref{theorem:p_softmax_complete} and \ref{theorem:p_softmax_minimal}, we prove the conditions under which $\Theta \subset \R^d$ is a complete and minimal representation of the Gumbel-Softmax probability distribution and generalize these results to other random perturbations, e.g., Gaussian-Softmax or more generally, Perturb-Softmax probability distributions.  We also extend these methods to Perturb-Argmax probability distributions, for which Gumbel-Argmax is a special case, and state the conditions for which it is complete and minimal in Theorems \ref{theorem:pert_max_complete} and \ref{theorem:p_argmax_minimal}. Our findings are illustrated in Figure \ref{fig:highlevel_properties}.
%The generality of our framework allows establishing these properties for any random perturbation, for example, Gaussian-Softmax or Gaussian-Argmax, or more broadly, Perturb-Softmax or Perturb-Argmax. 

%\tamir{paper structure}

We begin by introducing the notation relating parameters and the relevant probability distributions in Section \ref{sec:background}. Subsequently, we investigate the Perturb-Softmax probability models as gradients of the expected log-sum-exp convex function and prove their completeness by connecting their gradients to the relative interior of the probability simplex. In Section \ref{sec:perturbed_Softmax}, we determine the minimality of Perturb-Softmax by the strict convexity of the expected log-sum-exp when restricted to the respective parameter space. We then investigate the Perturb-Argmax probability models as sub-gradients of the expected-max convex function and establish the conditions for which their parameter space is complete and minimal, see Section \ref{sec:perturbed_argmax}. Finally, we empirically demonstrate the qualities of Perturb-Softmax extension in generative and discriminative learning setting, showing improved convergence of Gaussian-Softmax over Gumbel-Softmax beyond the linear high-dimensional setting that was investigated  in online learning \cite{abernethy2014online, pmlr-v37-cohena15}.

\section{Related work}
\label{sec:related}

The exponential family realized by the softmax operation over its parameters is extensively used in machine learning. However, sampling high-dimensional models is challenging due to its normalizing factor \citep{geman1984stochastic, goldberg2007complexity}. The Gumbel-Argmax probability distribution measures the stability of the argmax operation over Gumbel random variables. It serves as an equivalent representation of softmax operation, thereby enabling efficient sampling from the exponential family \citep{Gumbel1954, Luce59}. In the context of machine learning, the Gumbel-Argmax probability models underlie the "follow the perturbed-leader" online learning algorithm \citep{hannan1957approximation, kalai2002geometric, kalai2005efficient}. The Gaussian-Argmax probability models extend the "follow the perturbed-leader" family of online algorithms and improve their regret bounds \citep{NIPS2012_53adaf49, abernethy2014online, abernethy2016perturbation, pmlr-v37-cohena15}. Our work complements these studies by exploring the statistical properties of Gumbel-Argmax and Gaussian-Argmax. We prove the conditions under which Perturb-Argmax probability models are complete, making them suitable for use in machine learning, and when they are minimal, meaning they can uniquely identify a probability distribution from Perturb-Argmax probability models. 

The Gumbel-Softmax probability models were introduced as an alternative to the exponential family and its Gumbel-Argmax equivalent in the context of generative learning \citep{maddison2017,jang2017categorical}. This alternative allows for efficient sampling, making it highly effective for learning with stochastic gradient methods. The discrete nature of the Gumbel-Softmax sampling has been utilized to tokenize the visual vocabulary in the celebrated zero-shot text-to-image generation, DALL-E \citep{ramesh2021zero}. The Gaussian-Softmax probability models were introduced in variational auto-encoders as their closed-form KL-divergence makes it easier to realize as regularization \citep{NEURIPS2020_90c34175}. Our work extends these works and sets the statistical properties of Gumbel-Softmax and Gaussian-Softmax. We also show that Gaussian-Softmax enjoys faster convergence as the Gaussian distribution decays faster than the Gumbel distribution when approaching infinity. 

%The Gubeml-Argmax and Gumbel-Softmax stochastic gradient estimator has been successful applied to discriminative learning of high-dimensional discrete spaces, \citet{10.5555/3495724.3496202}. Its extension to learning bipartite graph matchings distributions allow to implicitly bias the learning process towards admissible matchings \citep{Gumbel-Sinkhorn_mena, indelman2023learning, indelman2024learning}. Further extensions were developed, such as sampling $k$ elements without replacement from a factorized distribution over sequences \citep{Kool2020AncestralGS, pmlr-v139-indelman21a}, subset sampling \cite{ErmonSubsetSamplig}, and more \citep{pmlr-v70-balog17a}. In reinforcement learning it is used to efficiently search for an optimal trajectory \cite{lorberbom2020direct}. In causality it was used for counterfactual reasoning \cite{oberst2019counterfactual, lorberbom2021learning}.  

\citet{48965} introduced a framework for optimizing discrete problems based on Perturb-Argmax probability models. This framework applies to discriminative learning using the Fenchel-Young losses and relies on convexity to propagate gradients over discrete choices. Similar to our approach, this work adopts a general view and utilizes convexity to explore the gradient properties of Perturb-Argmax models. Our work differs in that we use convexity and differentiability to investigate their statistical representation properties, specifically when these models are complete and minimal. Other methods include blackbox differentiation based on gradients of a surrogate linearized loss \citep{Vlastelica2020Differentiation}, the direct loss minimization technique \citep{NIPS2010_4069,Keshet2011DirectER,Song2016TrainingDN, pmlr-v139-indelman21a} based on gradients of the expected discrete 
loss, and entropy regularization techniques \citep{pmlr-v80-niculae18a, sparsemax}. Unlike these methods, we focus on studying the statistical properties of randomized discrete probability models rather than on optimization frameworks.

\section{Background}
\label{sec:background}

\iffalse In the following, we describe the completeness and minimality of the softmax operation to determine the statistical representation properties of a parameterized family of discrete probability distributions. We then describe the Gumbel-Softmax and Gumbel-Argmax and develop the statistical and functional framework to argue about their representation properties. \fi 

We denote by $\Delta$ the probability simplex, i.e., the set of all probabilities over $d$ discrete events, namely $\Delta \triangleq \{p \in \R^d : p(i) \ge 0, \sum_{i=1}^d p(i) = 1\}$. A parameterized discrete probability distribution $p_\theta(i) \in \Delta$ is determined by its parameters $\theta \in \Theta^d$ that reside in the Euclidean space $\Theta^d \subseteq \R^d$. 

\subsection{Completeness and minimality of the Softmax operation}\label{sec:sm}
The softmax operation $\softmax: \R^d \rightarrow \Delta$ is the standard mapping from the set of parameters $\Theta$ to the probability simplex $\Delta$. Formally, when we define $p^{sm}_\theta$ by softmax relation
\begin{equation}\label{eq:p_sm}
    p^{sm}_\theta \triangleq \softmax(\theta) \triangleq \left(\frac{e^{\theta_1}}{\sum_{j=1}^d e^{\theta_j}}, ..., \frac{e^{\theta_d}}{\sum_{j=1}^d e^{\theta_j}} \right).
\end{equation}

A parameterized family of distributions $\P_\Theta \triangleq \{ p_\theta: \theta \in \Theta\}$ is called complete if for every $p \in \Delta$ there exists $\theta \in \Theta$ such that $p_\theta = p$. Alternatively, the mapping from the parameters to their probabilities is onto the probability simplex (surjective). Similarly, a parameterized family of distributions is called minimal, if there is one-to-one mapping between its parameters and their corresponding probability distributions (injective). Formally, $p_\theta \ne p_\tau$ if and only if $\theta \ne \tau$. A complete and minimal mapping is also identifiable, i.e., for every probability $p \in \Delta$ one can identify the unique parameters $\theta \in \Theta$ for which $p = p_\theta$.  

The identifiability of the softmax mapping was explored in the context of the exponential family of distributions \citep{WainwrightJordan}. One can verify that the set of parameters $\Theta = \R^d$ is complete: for every $p \in \Delta$, one can set $\theta = \log p$, for which $p = softmax(\theta)$. However, the set $\Theta = \R^d$ is not minimal, as the two parameter vectors $\theta$ and $\theta + c 1$ both realize the same probability, i.e., $\softmax(\theta) = \softmax(\theta + c 1)$. Conversely, a set of parameters $\Theta$ is minimal for the softmax mapping if there are no $\theta, \tau \in \Theta$ for which $\theta_i = \tau_i = c$ for every $i =1, ..., d$.\footnote{If  $\softmax(\theta) = \softmax(\tau)$, and $p_\theta = \softmax(\theta)$, $p_\tau = \softmax(\tau)$, then $p_\theta(i) / p_\tau(i) = 1$ for every $i$, and $\log p_\theta(i) - \log p_\tau(i) = 0$. From the softmax mapping, this translates to $\theta_i - \tau_i = c$ for every $i$ while $c = \log(\sum_j e^{\theta_j} ) - \log(\sum_j e^{\tau_j} )$.} Consequently, identifiable sets of parameters for the softmax operation can be $\Theta = \{ \theta \in \R^d: \sum_j \theta_j = 0\}$ or $\Theta = \{ \theta \in \R^d:  \theta_1 = 0\}$. In Appendix \ref{theorem:Gibbs_minimal} we prove that these sets are both complete and minimal. 

\subsection{Gumbel-Softmax and Gumbel-Argmax probability distributions}

The Gumbel-Softmax probability distribution emerged as a smooth approximation of the Gumbel-Argmax representation of $p^{sm}_{\theta}$. We turn to describe the Gumbel-Argmax and Gumbel-Softmax discrete probability distributions.

The Gumbel distribution is a continuous distribution whose probability density function is $\hat g(t) = e^{-e^{-(t + c)}}$, where $c \approx 0.5772$ is the Euler-Mascheroni constant. We denote by $\gamma = (\gamma_1,...,\gamma_d)$ the vector of $d$ independent random variables that follow the Gumbel distribution law and by $g(\gamma) = \prod_{i=1}^d \hat g(\gamma_i)$ the probability density function (pdf) of the independent Gumbel distribution. 

We denote by $p^{gam}_\theta$ the Gumbel-Argmax probability distribution, which relies on a one-hot representation of the maximal argument. The indicator function $1[condition]$ equals one when the condition holds and zero otherwise. Then, the Gumbel-Argmax probability distribution takes the form:
\begin{eqnarray}\label{eq:p_theta_gam}
p^{gam}_\theta &\triangleq& \E_{\gamma \sim g} [\arg \max(\theta+\gamma)] \\
p^{gam}_\theta(i) &\triangleq& \E_{\gamma \sim g} [i = \arg \max(\theta+\gamma)]
\triangleq \int_{\R^d} g(\gamma) 1[i = \arg \max(\theta + \gamma)] d \gamma 
\end{eqnarray}

The fundamental theorem of extreme value statistics asserts the equivalence between the softmax distribution in Equation \ref{eq:p_theta_gsm} and the Gumbel-Argmax distribution in Equation \ref{eq:p_theta_gam}, namely $p^{sm}_\theta = p^{gam}_\theta$, cf. \cite{Gumbel1954}. Therefore, the statistical representation properties of completeness and minimality of the softmax operation are identical to the statistical properties of the Gumbel-Argmax probability distribution. Unfortunately, the argmax operation is non-smooth and requires special treatment when used in learning its parameters using gradient methods.

The Gumbel-Softmax probability distribution $p^{gsm}_\theta(i)$ was developed as a smooth approximation of its Gumbel-Argmax counterpart:   
\begin{eqnarray}
p^{gsm}_\theta &\triangleq& \int_{\R^d} g(\gamma) \softmax(\theta + \gamma) d \gamma 
\triangleq \E_{\gamma \sim g} [\softmax(\theta + \gamma)] \label{eq:p_theta_gsm}
\end{eqnarray}
We note that we define $p^{gsm}_\theta$ as a $d$-th dimensional vector, and the integral with respect to $\gamma$ (or expectation) is taken with respect to each coordinate of the softmax operation.

One can verify that the Gumbel-Softmax is indeed a probability distribution: (i) it is non-negative since it is the average of the non-negative softmax operation, and (ii) it sums up to unity since it is an average of the softmax operations $\softmax(\theta + \gamma)$ that sum up to unity for every $\gamma$. 

\iffalse In this work, we explore the statistical representation properties of the Gumbel-Sofmax operation. In Theorem XXX We prove that $\Theta \subset \R^d$ is complete and minimal whenever XXX. As a consequence of our technique, we prove that any random perturbation, e.g., Gaussian-Softmax or Discrete-Softmax, is complete and minimal under the same conditions. We also apply these methods to Gumbel-Argmax probability models and state the conditions for which it is complete and minimal in Theorem YYY. Our statistical investigation relies on the differential properties of convex functions.  \fi

\subsection{Differentibility properties of convex functions}

We investigate the representation properties of the Gumbel-Softmax and Gumbel-Argmax probability models when they result from gradients of multivariate functions over their set of parameters $\Theta$. 

The softmax function is the gradient of the log-sum-exp function: 
\begin{equation}
    \softmax(\theta) = \nabla \log \left( \sum_{i=1}^d e^{\theta_i} \right) \label{eq:lse_grad}.
\end{equation}
As shown in Section \ref{sec:sm} this gradient mapping is complete and minimal over a convex subsetset $\Theta \subset \R^d$. The extension of this argument to Gumbel-Softmax requires notions of convexity, covered in Appendix \ref{app:related_convexity}. 
The (sub)differential of different convex functions $f(\theta)$ is used in our study as a (multi-valued) mapping between the convex set of the primal domain $\Theta$ and its dual domain $\calP$, which is the Gumbel-Softmax or the Gumbel-Argmax probability model.
We define the conditions for which:
\begin{enumerate}[noitemsep,topsep=0pt,label=\Roman*]
    \item $\partial f$ is a single-valued mapping, i.e., $\partial f = \nabla f$. In this case, for every $\theta \in \Theta$ matches a single $\nabla f(\theta) \in \calP$. If this property does not hold then the parameters $\theta$ that generate a probability $p$ are not identifiable under this mapping. 
    \item The gradient mapping $\nabla f(\theta)$ is onto the probability simplex $\Delta$. In this case, the set of parameters $\Theta$ is complete, i.e., it can represent (and learn) any probability $p \in \Delta$ using its gradients $\nabla f(\theta)$.  
    \item The gradient mapping $\nabla f(\theta)$ is one-to-one. In this case, the set of parameters $\Theta$ is minimal, i.e., there are no two parameters $\theta, \tau$ that represent the same probability distribution $p \in \calP$. 
\end{enumerate}
Our framework allows establishing these relations for any random perturbations, which we refer to as Perturb-Softmax and Perturb-Argmax. %This provides theoretical justification to the empirical observation that for fitting probabilities one need not limit oneself to Gumbel random perturbations. 

\section{Perturb-Softmax probability distributions}\label{sec:perturbed_Softmax}
In this section, we explore the statistical representation properties of the Perturb-Softmax operation as a generalization of the Gumbel-Softmax operation. 
Our exploration emerges from the connection between the softmax operation and the log-sum-exp convex function, as described in Equation {\ref{eq:lse_grad}}. We establish a similar relation between perturb-log-sum-exp and perturb-softmax: 
\begin{eqnarray}
f(\theta) &=& \E_{\gamma} \left[ \log \left( \sum_{i=1}^{d} e^{\theta_i+\gamma_i} \right) \right] \label{f_theta_logsumexp_perturbed}\\
\nabla f(\theta) &=& \E_{\gamma} [\softmax(\theta + \gamma)] \label{eq:psm}
\end{eqnarray}
The function $f(\theta)$ is defined for any random perturbation $\gamma$, whether $\gamma$ values are from a discrete, a bounded, or an unbounded set \footnote{Formally, for unbounded random perturbations $\gamma$ we restrict ourselves to probability density functions for which $f(\theta) < \infty$.}. The function $f(\theta)$ is differentiable since it is the expectation of the differentiable log-sum-exp function and $\nabla f(\theta)$ is attained by the Leibniz rule for differentiation under the integral sign\footnote{Formally, $\nabla f(\theta)$ is finite whenever the dominant convergence theorem holds. For unbounded $\gamma$ this holds for any probability density function $p(\gamma)$ for which $\lim_{\gamma \rightarrow \infty} p(\gamma) \log \left( \sum_{i=1}^{d} e^{\theta_i+\gamma_i} \right) = 0$. This happens for Gumbel, Gaussian, and other standard probability density functions.}. Also, the function $f(\theta)$ is convex, as it is an expectation of  convex log-sum-exp functions. We exploit the convexity of $f(\theta)$ to define the conditions on the parameter space $\Theta$ for which the Perturb-Softmax probability distributions span the probability simplex.   

The gradient $\nabla f(\theta)$ maps parameters $\theta$ to a probability, as the softmax vector $\softmax(\theta + \gamma)$ sums up to unity for any $\gamma$ and therefore also in expectation over $\gamma$ (Corollary \ref{cor:logsumexp_derivative_probability_func}). In the next theorem, we determine the conditions for which the gradient mapping spans the relative interior of the probability simplex, i.e., the set of all possible positive probabilities. 

\begin{theorem}[Completeness of Perturb-Softmax]\label{theorem:p_softmax_complete}
    Let $\Theta \subseteq \R^d$ be a convex set and let $\gamma = (\gamma_1,...,\gamma_d)$ be a vector of random variables whose cumulative distribution decays to zero as $\gamma$ approaches $\pm \infty$. Let $h_i(\theta) = \theta_i - max_{j \ne i} \theta_j$ be a continuous function over $\Theta$. If $h_i(\theta)$ are unbounded then $\Theta$ is a complete representation of the Perturb-Softmax probability models:   
    \begin{equation}
        ri(\Delta) \subseteq \E_{\gamma} [\softmax(\theta + \gamma)] \subseteq \Delta  
    \end{equation}
\end{theorem}
\begin{proof}
The proof relies on fundamental notions of the conjugate dual function $f^*(p)$ and its convex domain $\calP$, cf. Equations (\ref{eq:fstar}, \ref{eq:domfstar}) in the Appendix. 

We begin by considering $f(\theta) = \E_{\gamma} [ \log ( \sum_{i=1}^{d} e^{\theta_i+\gamma_i} ) ]$ and its gradient, which is the Perturb-Softmax model $\nabla f(\theta) = \E_{\gamma} [\softmax(\theta + \gamma)]$. Equation \ref{eq:rockaffeller_rangeproof} implies that its gradients, i.e., the Perturb-Softmax probability distributions,  reside in their convex domain $\calP$ (cf. Equation \ref{eq:partialf}), and contain its relative interior, thus: 
\begin{equation}
    ri(\calP) \subseteq \left\{ \E_{\gamma} [\softmax(\theta + \gamma)]: \theta \in \Theta 
    \right\} \subseteq \calP  
\end{equation}
To conclude the proof, we prove in Appendix \ref{app:completeness_perturbsoftmax} that the zero-one probability vectors reside in the closure of $\cal P$. Since the closure of $\cal P$ is a convex set, we conclude that it is the probability simplex. 
\iffalse
Given the conditions of the theorem on $h_i(\theta)$, we can construct a series $\{ \theta^{(n)} \}_{n=1}^\infty$ for which $h_i(\theta^{(n)}) = n$ for every $n \in \mathbb{N}$. We show that  $\E_{\gamma} [\arg \max (\theta^{(n)} + \gamma)]$ approaches the zero-one probability vector as $n \rightarrow \infty$. $\E_\gamma \left[\frac{e^{\theta^{(n)}_i + \gamma_i}}{\sum_{j=1}^d e^{\theta^{(n)}_j + \gamma_j}} \right] =$
\begin{eqnarray}
     \E_\gamma \left[\frac{1}{\sum_{j=1}^d e^{\theta^{(n)}_j + \gamma_j - \theta^{(n)}_i - \gamma_i}} \right]
    = \E_\gamma \left[ \frac{1}{1 + \sum_{j \ne i} e^{\theta^{(n)}_j - \theta^{(n)}_i + \gamma_j - \gamma_i}} \right] 
    \ge \E_\gamma \left[ \frac{1}{1 + \sum_{j \ne i} e^{-n + \gamma_j - \gamma_i}} \right]  \stackrel{n \rightarrow \infty}{\rightarrow 1}
\end{eqnarray}
The limit argument holds since the probability of $\gamma_1,...,\gamma_d$ decay as they tend to infinity. This proves that the zero-one distributions are limit points of probabilities in $\calP$, i.e., $cl(\calP) = \Delta$. \fi
\end{proof}
The above theorem implies that the set of all Perturb-Softmax probability distribution is an almost convex set that resides within the convex set of all probabilities $\Delta$ and contains its relative interior, i.e., the convex set of all positive probabilities $ri(\Delta)$.

Next, we describe the conditions for which the parameter space $\Theta$ is minimal. In this case, two different parameters $\theta \ne \tau$ result in two different Perturb-Softmax models $\E_{\gamma} [\softmax(\theta + \gamma)] \ne \E_{\gamma} [\softmax(\tau + \gamma)]$. Interestingly, minimality is tightly tied to strict convexity. We begin by proving that $f(\theta) = \E_{\gamma} [ \log ( \sum_{i=1}^{d} e^{\theta_i+\gamma_i} ) ]$ is strictly convex when restricted to $\Theta$. 

\begin{lemma}[Strict convexity] \label{lemma:strictconvexity_softmax}
    Let $\Theta \subseteq \R^d$ be a convex set and let $\gamma = (\gamma_1,...,\gamma_d)$ be a vector of random variables and let $f(\theta) = \E_{\gamma} [ \log ( \sum_{i=1}^{d} e^{\theta_i+\gamma_i} ) ]$. If 
    $\Theta$ has no two vectors $\theta \ne \tau \in \Theta$ that are affine translations of each other,  for which $\theta_i = \tau_i + c$ for every $i=1,...,d$ and some constant $c$ then $f(\theta)$ is strictly convex over $\Theta$, i.e., for any $\theta \ne \tau \in \Theta$ and any $0 < \lambda < 1$ it holds that
\begin{equation}
f(\lambda \theta + (1- \lambda)\tau ) < \lambda f(\theta) + (1-\lambda) f(\tau)    .
\end{equation}
\end{lemma}
The proof, based on H\"{o}lder's inequality, is provided in Appendix \ref{app:strict_convexity_e_log_sum_exp_proof}.
\iffalse
\begin{proof}
Let $\theta, \tau \in \R^d$ and $0 < \lambda < 1$. Then 
\begin{eqnarray}
    f(\lambda \theta + (1- \lambda)\tau )= \E_\gamma \left[ \log \left( \sum_{i=1}^{d} e^{\lambda (\theta_i + \gamma_i) +(1- \lambda)(\tau_i + \gamma_i) } \right) \right] 
    = \E_\gamma \left[ \log \left(\sum_{i=1}^{d} u_i v_i \right) \right], \label{eq:perturb_log_sum_eq_dim}
\end{eqnarray} 
where $u_i \triangleq e^{\lambda (\theta_i + \gamma_i)}$ and $v_i \triangleq e^{(1- \lambda)(\tau_i + \gamma_i) }$. Applying H\"{o}lder's inequality $\langle u,v \rangle \le \| v \|_{1/\lambda} \cdot \| u\|_{1/(1-\lambda)}$ we obtain the convexity condition of the log-sum-exp function: 
\begin{equation}
f(\lambda \theta + (1- \lambda)\tau ) \le \lambda f(\theta) + (1-\lambda) f(\tau).
\end{equation}
To prove strict convexity we note that H\"{o}lder's inequality for non-negative vectors $u_i, v_i$ holds with equality if and only if there exists a constant $\alpha \in \R$ such that $v_i = \alpha u_i^{\frac{1 - \lambda}{\lambda}}$ for every $i=1,...,d$, or equivalently:
\begin{equation}
        e^{(1- \lambda)(\tau_i + \gamma_i)} = \left( e^{\lambda (\theta_i + \gamma_i)}\right)^{\frac{1 - \lambda}{\lambda}} \longleftrightarrow \tau_i = \theta_i + c
\end{equation}
Where $c = \frac{\log \alpha}{1-\lambda}$. Therefore, if $\tau \ne \theta + c$, then $\langle u,v \rangle < \| v \|_{1/\lambda} \cdot \| u\|_{1/(1-\lambda)}$ for every $\gamma$ and consequently it also holds when applying the logarithm function and taking an expectation with respect to $\gamma$. 
\end{proof}
\fi
The condition that $\theta$ and $\tau$ are not a translation of each other guarantees strict convexity. If $\tau_i = \theta_i + c$ for every $i$, then $f(\tau) = f(\theta) + c$. This linear relation implies that the convexity condition holds with equality.\iffalse: $f(\lambda \theta + (1-\lambda) \tau) = \lambda f(\theta) + (1-\lambda) f(\tau)$. \fi

The minimality theorem is a direct consequence of Lemma \ref{lemma:strictconvexity_softmax}, as strict convexity of differentiable function implies the gradient mapping is one-to-one (cf. \citet{rockafellar-1970a}, Theorem 26.1).  

\begin{theorem}[Minimality of Perturb-Softmax] \label{theorem:p_softmax_minimal}
    Let $\Theta \subseteq \R^d$ be a convex set and let $\gamma = (\gamma_1,...,\gamma_d)$ be a vector of random variables. $\Theta$ is a minimal representation of the Perturb-Softmax probability models if there are no two parameter vectors $\theta \ne \tau \in \Theta$ that are affine translations of each other, for which $\theta_i = \tau_i + c$ for every $i=1,...,d$ and some constant $c$.    
\end{theorem}
\begin{proof}
    Lemma \ref{lemma:strictconvexity_softmax} implies that $f(\theta)$ is strictly convex and Equation (\ref{eq:psm}) implies it is differentiable.  Recall the conjugate dual function and its domain (Equations (\ref{eq:fstar}, \ref{eq:domfstar}) in the Appendix) and its gradient mapping $\nabla f: \Theta \rightarrow \calP$. Since $f(\theta)$ is strictly convex then the function $g(\theta) = \langle p, \theta \rangle - f(\theta)$ is strictly concave hence its maximal argument $\theta^*$ is unique. The gradient vanishes at the maximal argument, $\nabla g(\theta^*) = 0$ or equivalently, $ p = \nabla f(\theta^*)$. Since $\theta^*$ is unique then $p$ is unique as well. Therefore $\nabla f(\theta)$ is a one-to-one mapping.  
\end{proof}

To conclude, one can use any convex set $\Theta \subset \R^d$ that satisfies the conditions of Theorem \ref{theorem:p_softmax_complete} and Theorem \ref{theorem:p_softmax_minimal}. Similarly to the softmax probability model, $\Theta$ that is both complete is minimal can be $\Theta = \{ \theta \in \R^d: \sum_j \theta_j = 0\}$ or $\Theta = \{ \theta \in \R^d:  \theta_1 = 0\}$.

\section{Perturb-Argmax probability distributions}\label{sec:perturbed_argmax}

In this section, we explore the statistical representation properties of the Perturb-Argmax operation as a generalization of the Gumbel-Argmax operation. Throughout our investigation, we treat the Perturb-Argmax probability model as the sub-gradient of the expected Perturb-Max function, which we prove in Corollary \ref{cor:derive_probability_max} in the Appendix for completeness:
\begin{eqnarray}
f(\theta) &=& \E_{\gamma} \left[ \max_{i} \{ \theta_i+ \gamma_i\}  \right] \label{g_theta_max_perturbed}\\
\partial f(\theta) &=& \E_{\gamma} [\arg \max(\theta+\gamma)]\label{eq:derivative_expected_max}
\end{eqnarray}
Different than the softmax operation, the argmax operation is not continuous everywhere. This difference arises from the fact that unlike the differentiable log-sum-exp function, the max function is not everywhere differentiable. However, since it is a convex function, its sub-gradient always exists. 
In the following, we prove that the sub-gradients span the set of all positive probability distributions, i.e., the relative interior of the probability simplex. 
\begin{theorem}[Completeness of Perturb-Argmax]\label{theorem:pert_max_complete}
    Let $\Theta \subseteq \R^d$ be a convex set and let $\gamma = (\gamma_1,...,\gamma_d)$ be a vector of random variables. Then, $\Theta$ is a complete representation of the Perturb-Argmax probability models:   
    \begin{equation}
        ri(\Delta) \subseteq \E_{\gamma} [
        \arg\max(\theta + \gamma)] \subseteq \Delta  
    \end{equation}
    
\end{theorem}

\begin{proof}
    
The proof technique follows the argument of Theorem \ref{theorem:p_softmax_complete}. Given the conditions on $h_i(\theta)$, we can construct a series $\{ \theta^{(n)} \}_{n=1}^\infty$ for which $h_i(\theta^{(n)}) = n$ for every $n \in \mathbb{N}$. To conclude the proof, we prove in Appendix \ref{app:completeness_perturbargmax} that $\E_{\gamma} [\arg \max(\theta^{(n)} + \gamma)]$ approaches the zero-one probability vector as $n \rightarrow \infty$.
\iffalse
Given the conditions on $h_i(\theta)$, we can construct a series $\{ \theta^{(n)} \}_{n=1}^\infty$ for which $h_i(\theta^{(n)}) = n$ for every $n \in \mathbb{N}$.  We show that $\E_{\gamma} [\arg \max(\theta^{(n)} + \gamma)]$ approaches the zero-one probability vector as $n \rightarrow \infty$. 
\begin{eqnarray}
     \P[i = \arg \max(\theta^{(n)} + \gamma) & =& \P \left[ \theta^{(n)}_i + \gamma_i \ge \max_{j \ne i} \{\theta^{(n)}_j + \gamma_j\} \right]  \\
    & \ge & \P \left[\theta^{(n)}_i + \gamma_i \ge \max_{j \ne i} \{\theta^{(n)}_j\} + \max_{j \ne i} \{ \gamma_j \}  \right] 
    \ge \P \left[\gamma_i \ge -n + \max_{j \ne i} \{ \gamma_j \} \right] \stackrel{n \rightarrow \infty}{\rightarrow 1} \nonumber
\end{eqnarray}
The limit argument holds since the probability of $\gamma_1,...,\gamma_d$ decay as they tend to infinity.\fi  This proves that the zero-one distributions are limit points of probabilities in $\calP$, i.e., $cl(\calP) = \Delta$. 
\end{proof}

The above theorem holds for any type of random perturbation $\gamma$. Next, we show that the statistical properties of the Perturb-Argmax probability models depend on their perturbation type. 
The minimality of the representation of Perturb-Argmax probability models holds for non-discrete random perturbation $\gamma$. It relies on the differentiability properties of its probability density function $p(\gamma)$. 
\begin{lemma}[Differentiability of Perturb-Max] \label{lemma:diffofPerturb-Max}
    Let $\gamma = (\gamma_1,...,\gamma_d)$ be a vector of random variables with differentiable probability density function $p(\gamma) = \prod_{i=1}^d p_i(\gamma_i)$ and let $f(\theta) = \E_{\gamma} [ \max \{\theta + \gamma \}]$. Then, $f(\theta)$ is differentiable and its gradient is 
    \begin{equation}
        \nabla f(\theta) = \E_{\gamma} [\arg \max(\theta+\gamma)]        
    \end{equation}
\end{lemma}
The proof is provided in Appendix \ref{app:diffofPerturb-Max}.
Lemma \ref{lemma:diffofPerturb-Max} shows a single-valued mapping from the parameter space to the probability space. In the following, we show that this mapping brings forth a minimal representation of the Perturb-Argmax probability models under certain conditions. 

\begin{theorem}[Minimality of Perturb-Argmax]\label{theorem:p_argmax_minimal}
    Let $\Theta \subseteq \R^d$ be a convex set and let $\gamma = (\gamma_1,...,\gamma_d)$ be a vector of random variables whose probability density functions $p_i(\gamma_i)$ are differentiable and positive. $\Theta$ is a minimal representation of the Perturb-Argmax probability models if there are no two parameter vectors $\theta \ne \tau \in \Theta$ that are affine translations of each other, for which $\theta_i = \tau_i + c$ for every $i=1,...,d$ and some constant $c$.    
\end{theorem}
The proof is provided in Appendix \ref{app:minnimality_perturb_argmax}. It is based on showing that under these conditions, the function $f(\theta) = \E_\gamma [\max \{\theta + \gamma\}]$ is strictly convex. We rely on its one-dimensional function $g(\lambda) \triangleq f(\theta + \lambda v)$ and show that $g''(\lambda) > 0$. Since the function $g(\lambda)$ is convex then $g''(\lambda) \ge 0$, and it is enough to show that $g'(\lambda)$ depends on $\lambda$, for which it follows that $g''(\lambda) \ne 0$ and consequently $g''(\lambda) > 0$. 

Our theorem conditions require the probability density function to be positive, to ensure that the second derivative is positive as it always accounts for a change in the perturbation space. 

\subsection{Non-minimal representation for bounded perturbations }\label{sec:max_smooth}

In the following, we analyze an example of a Perturb-Argmax distribution when the probability density function of the perturbation is differentiable almost everywhere but bounded. In this case, one can construct a non-minimal representation.

\begin{proposition}\label{prop:nonminimal_argmax_bounded}
     Let $\theta \in \R^2$, and consider i.i.d. random variables with a smooth bounded probability density function $\gamma \sim U(1,-1)$. A single-valued mapping exists between $f(\theta)$ and the Perturb-Argmax probability distribution. However, a one-to-one mapping does not exist.
 \end{proposition}
\begin{proof}

The perturb-max function $f(\theta)$ can be expressed as
\begin{align}
    f(\theta) =\theta_2 + \E_{\gamma_1,\gamma_2} \left[ \max\{ \theta_1-\theta_2 + \gamma_1 -\gamma_2 ,0 \}  \right], 
\end{align}
when the distribution of $\gamma$ is omitted for brevity.
\iffalse Define $\theta = \theta_1-\theta_2$ and $Z = \gamma_1-\gamma_2$. Then, the random variable $Z$ has a triangular distribution, and the following probability distribution function:
\begin{eqnarray}\label{pdf_smoothbounded_}
    f_{Z}(z) &=& 
    \begin{cases}
        \frac{1}{4}(2+z) \quad\text{if } 0 > z \geq -2\\
        \frac{1}{4}(2-z) \quad\text{if } 2 \geq z \geq 0 \\ 
         0 \quad\text{otherwise }.
    \end{cases}
\end{eqnarray}
\fi
We can express $f(\theta)$ by the pdf of the random variable $\gamma_1-\gamma_2$ (Equation \ref{ftheta_boundedsmooth_} in the appendix). Since $f(\theta)$ is a smooth function, a single-valued mapping exists (Theorem 26.1 \cite{rockafellar-1970a}). However, $f(\theta)$ is not strictly convex, hence a one-to-one mapping does not exist and it can be concluded that $\Theta$ is not a minimal representation of the Perturb-Argmax probability.
\iffalse
Then, the derivatives of $f(\theta)$ w.r.t. $\theta$ corresponding to the probabilities of the $\arg \max$ take the values:
\begin{align}
    \frac{\partial}{\partial \theta}f(\theta) &=
    \begin{cases}
        (1,0) \quad\text{if }  \theta > 2 \\ 
        (\frac{1}{2}+\frac{1}{2}\theta-\frac{1}{8}\theta^2,\frac{1}{2}-\frac{1}{2}\theta + \frac{1}{8}\theta^2 ) \quad\text{if } 2 \geq \theta \geq 0 \\ 
        (\frac{1}{2}+ \frac{1}{2}\theta+ \frac{1}{8}\theta^2, \frac{1}{2}- \frac{1}{2}\theta - \frac{1}{8}\theta^2) \quad\text{if } 0 \geq \theta \geq -2 \\
       (0,1)\quad\text{if }  \theta < -2.
\end{cases}
\end{align} 
\fi
The derivatives of $f(\theta)$, corresponding to the probabilities of the $\arg \max$, are illustrated in Figure \ref{fig:ftheta_derivatives_boundedsmooth_theta_ranges_plot}.

\begin{figure*}[t]
  \centering
    \begin{minipage}[t]{0.48\textwidth}
    \centering
    \includegraphics[width=0.75\textwidth]{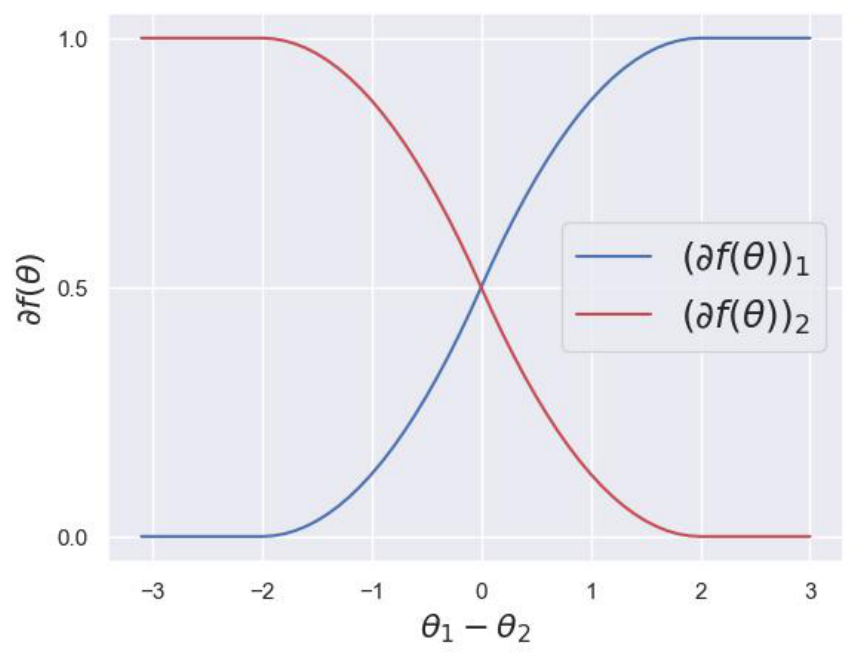}
    \caption{An illustration of $\partial f(\theta)$ for perturbations with a smooth bounded probability density function $\gamma \sim U(-1,1)$. $\partial f(\theta)$ is a single-valued mapping between the parameters and the Perturb-Argmax probability. }
\label{fig:ftheta_derivatives_boundedsmooth_theta_ranges_plot}
\end{minipage}\hfill
\begin{minipage}[t]{0.49\textwidth}
    \centering
\includegraphics[width=0.76\textwidth]{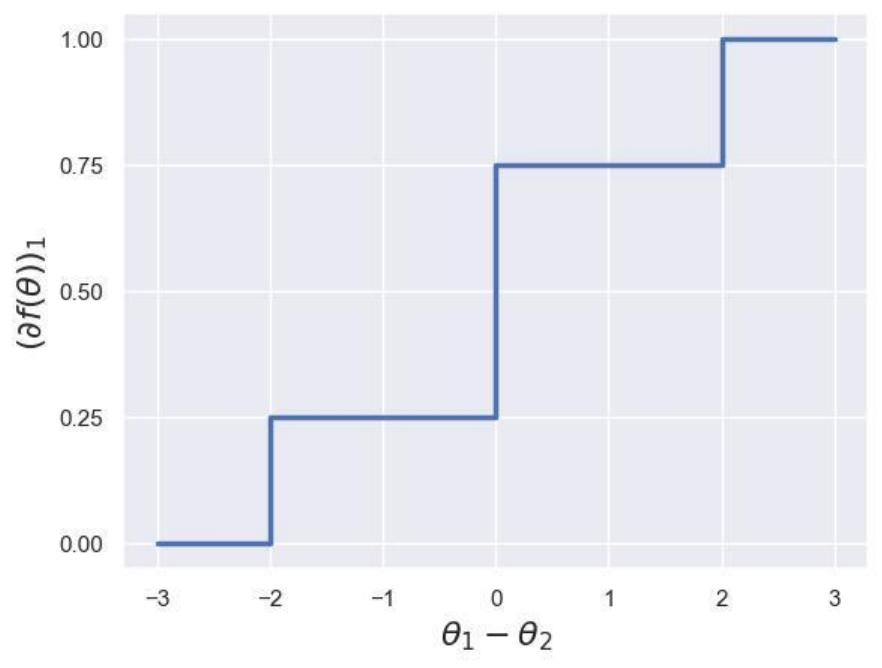}
\caption{An illustration of the sub-differential of $f(\theta)$ (Equation \ref{discrete_max_ranges}) w.r.t. $\theta_1$ for discrete random variables $\gamma_i \in \{1, -1\}$ that are uniformly distributed. Notably, the Perturb-Argmax probability is a multi-valued mapping in its overlapping segments, e.g., for $\theta_1= \theta_2$,}
    \label{fig:map_prob_illustr}
\end{minipage}
\end{figure*}

We defer all details to Section \ref{app:max_sbounded} in the Appendix.
 \end{proof}

\subsection{Discrete perturbations and identifiablity}\label{sec:max_discrete}

We next analyze the identifiability of the Perturb-Argmax distribution representation. Since the max function is not always differentiable, the Perturb-Max function $f(\theta)$ (Equation \ref{g_theta_max_perturbed}) is not always differentiable. However, since the max function is convex, its sub-differential $\partial f(\theta)$ exists. Unfortunately, the function $\partial f(\theta)$ is a multi-valued function, i.e., for some parameters $\theta \in \Theta$ there exist $p \ne q \in \calP$ that both are its sub-gradient. Thus, the probability cannot be identified from the parameters when the perturbation is discrete. This property is demonstrated in the following proposition: 
\begin{proposition}\label{prop:max_discrete}
Let $\Theta = \R^2$ and $\gamma = (\gamma_1,\gamma_2)$ be a vector of discrete random variables $\gamma_i \in \{1, -1\}$ that are uniformly distributed: $\P[\gamma_i = 1] = \P[\gamma_i = -1] = \frac{1}{2}$. Then, the Perturb-Argmax probability distribution is not identifiable.
\end{proposition}

The proof is provided in Appendix \ref{app:h_max_discrete}, and it is based on computing the function $f(\theta) = \E_\gamma [\max\{\theta + \gamma\}]$ analytically by taking the expectation w.r.t. $\gamma$.
The function $f(\theta)$, illustrated in Figure \ref{app:subgradient_illustr} in the appendix, is continuous and differentiable almost everywhere. 
However, in its overlapping segments\iffalse , i.e., when $\theta_1 = \theta_2 + 2$, $\theta_1 = \theta_2$ and $\theta_1 + 2 = \theta_2$\fi, the function is not differentiable, i.e., it has a sub-differential $\partial f(\theta)$ which is a set of sub-gradients. To prove that the Perturb-Argmax probability model is unidentifiable, we show that $\partial f(\theta)$ is a multi-valued mapping when $\theta_1 = \theta_2$. \iffalse In particular, we show that every probability distribution $p = (p_1,p_2)$ with $p_1 \in [\frac{1}{4}, \frac{3}{4}]$ satisfies $p \in \partial f(\theta)$. \fi The sub-differential mapping is illustrated in Figure \ref{fig:map_prob_illustr}.

\section{Experiments}
\label{sec:experiments}
In this section, we demonstrate the advantage of the Gaussian-Softmax over the commonly used Gumbel-Softmax. Experiments in density estimation and variational inference exhibit that, compared to the Gumbel-Softmax, the Gaussian-Softmax enjoys a faster convergence rate and better approximate discrete distributions.

\subsection{Approximating discrete  distributions}
\begin{figure}
\centering
\begin{subfigure}[c]{\textwidth}
  \centering
  \includegraphics[width=0.29\linewidth]{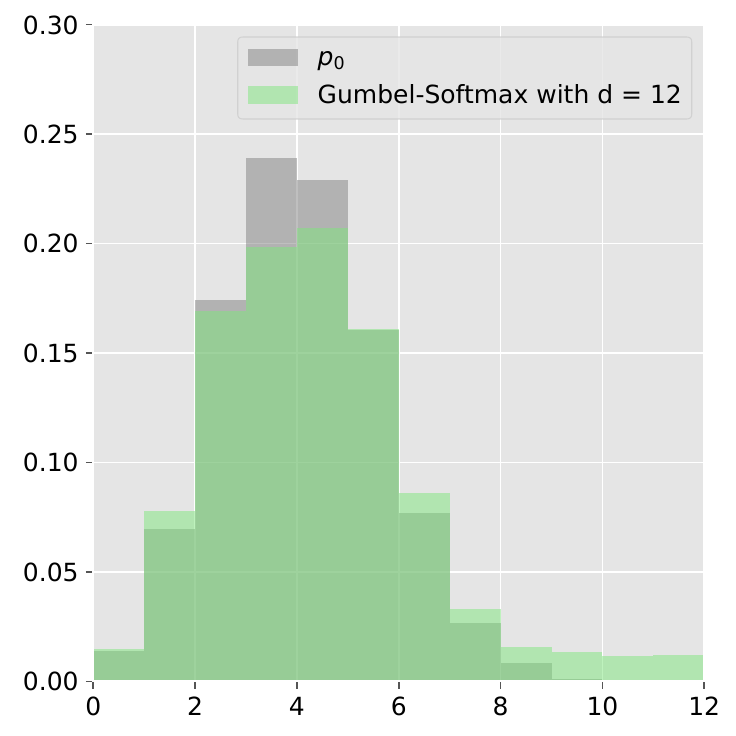}
  \includegraphics[width=0.29\linewidth]{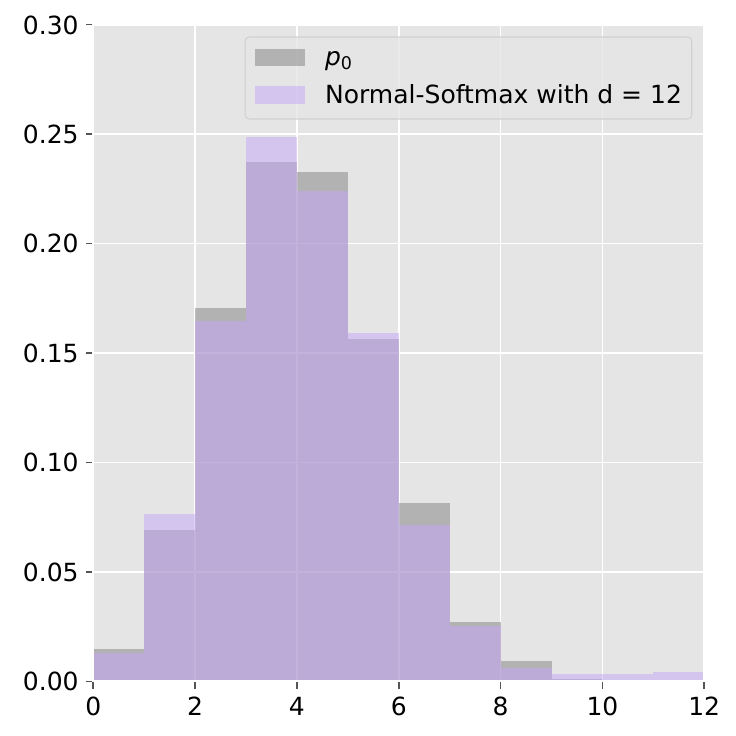}
  \includegraphics[width=0.33\linewidth]{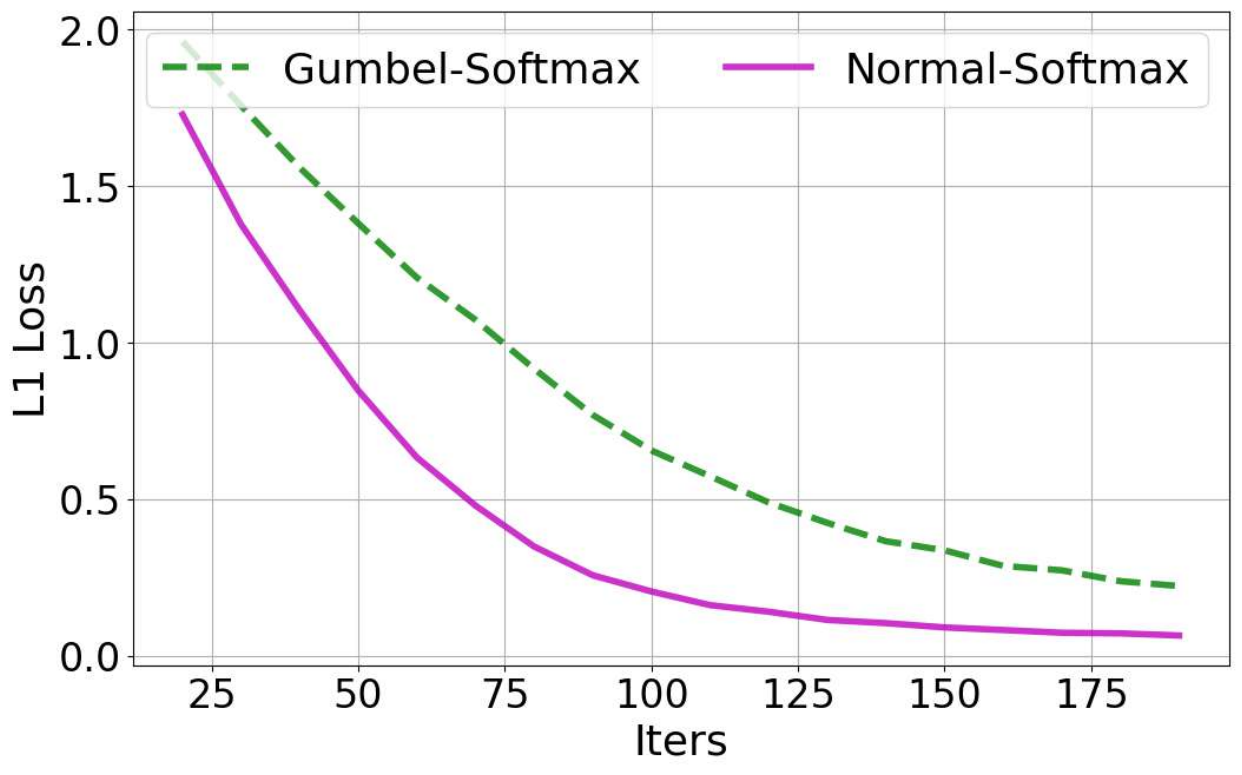}
  \caption{Approximation with dimension d=12 of a binomial distribution.}
\label{fig:Case2_binomial}
\end{subfigure}
\vskip\baselineskip
\begin{subfigure}[c]{\textwidth}
  \centering
  \includegraphics[width=0.29\linewidth]{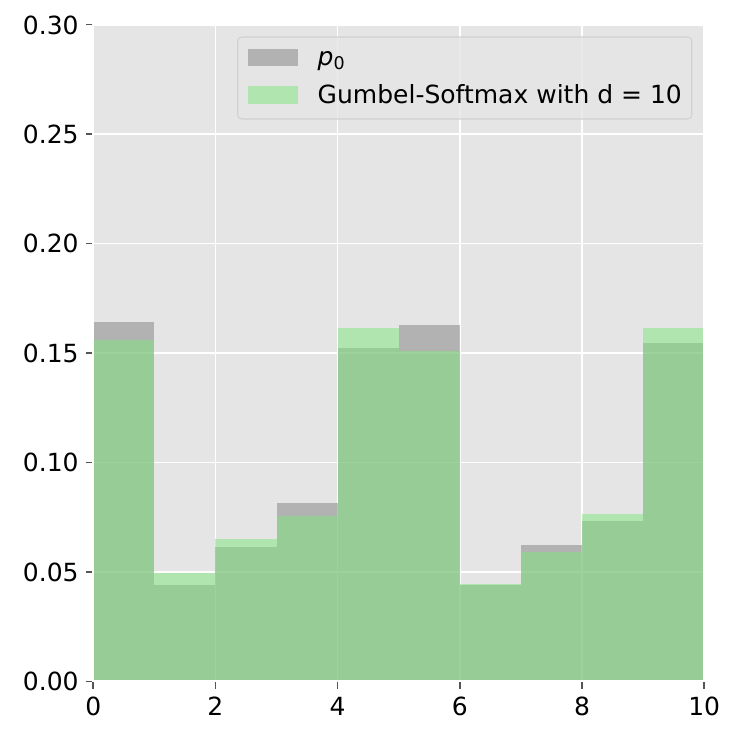}
  \includegraphics[width=0.29\linewidth]{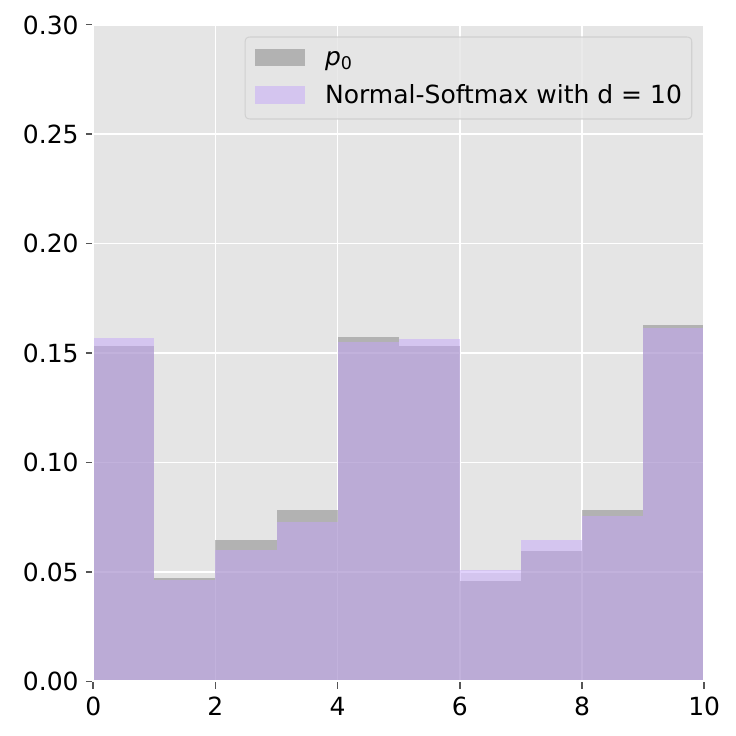}
  \includegraphics[width=0.33\linewidth]{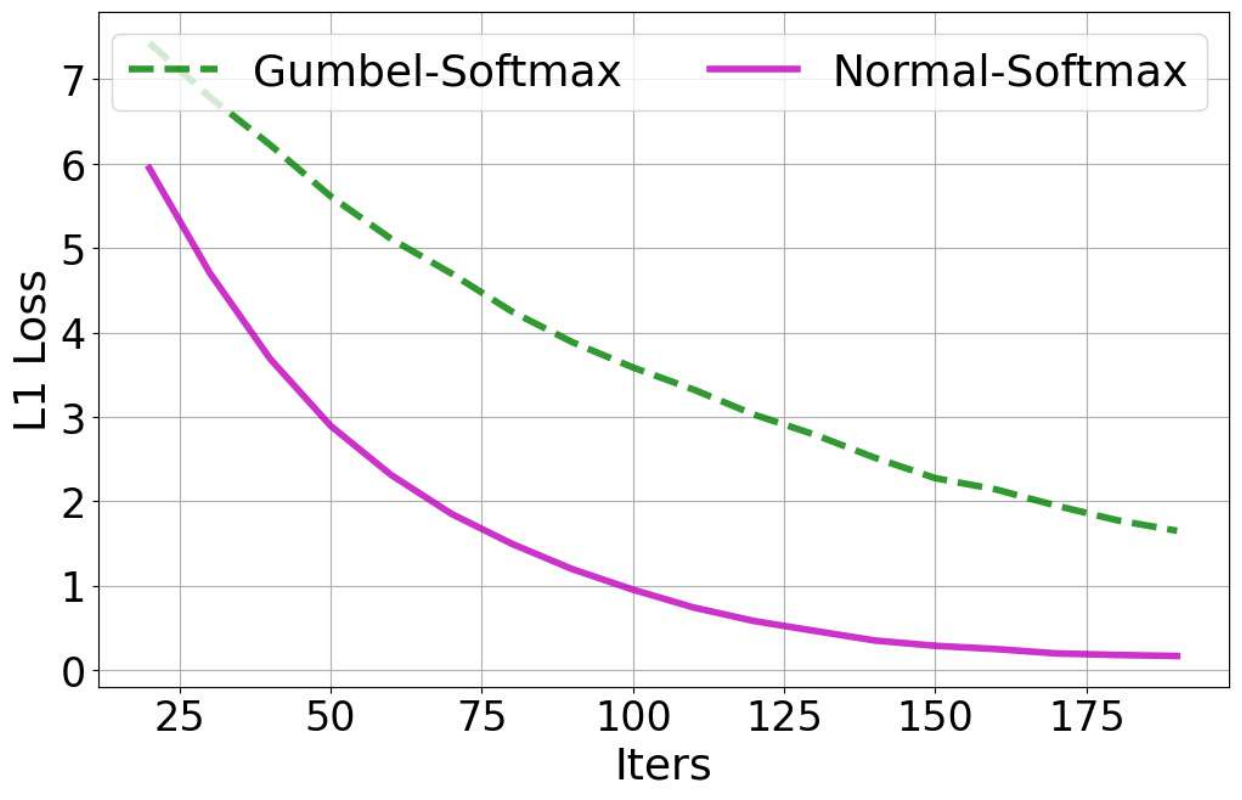}
  \caption{Approximation with dimension d=10 of a discrete distribution.}
  \label{fig:Case1_discrete}
\end{subfigure}
\caption{Gumbel-Softmax and Normal-Softmax approximation of target discrete distributions $p_0$ with finite support. The $L1$ objective over learning iterations is depicted on the right.}
\label{fig:approx_finitesupport}
\end{figure}

We compare the Gumbel-Softmax and the Normal-Softmax in approximating discrete distributions. The $L_1$ objective function is minimized between the Perturb-Softmax function applied to the fitted parameters of the probability density function and the target discrete distribution, denoted by $p_0$. We consider two target discrete distributions with finite support: a binomial distribution with parameters $n=12,p=0.3$, and a discrete distribution with $p = (\frac{10}{68},\frac{3}{68}, \frac{4}{68}, \frac{5}{68}, \frac{10}{68}, \frac{10}{68}, \frac{3}{68}, \frac{4}{68}, \frac{5}{68}, \frac{10}{68})$. Figure \ref{fig:approx_finitesupport} shows that the Normal-Softmax better approximates both distributions and exhibits faster convergence than the Gaussian-Softmax. 

Following the experiment in \cite{NEURIPS2020_90c34175}, we consider discrete distributions with countably
infinite support: a Poisson distribution with $\lambda=50$,  and a negative binomial distribution with $r=50, p=0.6$.
Differently from our method, the identifiability of the parameters is lost with the invertible Gaussian parameterization method. Results show that the Normal-Softmax has similar benefits over the Gaussian-Softmax as exhibited for discrete distributions with finite support (Figure \ref{fig:approx_dist_infinitesupport} in the appendix). 
Table \ref{table:aproxx-dist-l1} in the Appendix shows that the approximation based on the Normal-Softmax probability model consistently achieves lower mean and standard deviation $L_1$.
See more details in Appendix \ref{app_exp_apprpx}.

\subsection{Variational inference}
\begin{figure}[h]
\centering
\begin{subfigure}[b]{.32\textwidth}
\includegraphics[width=\linewidth]{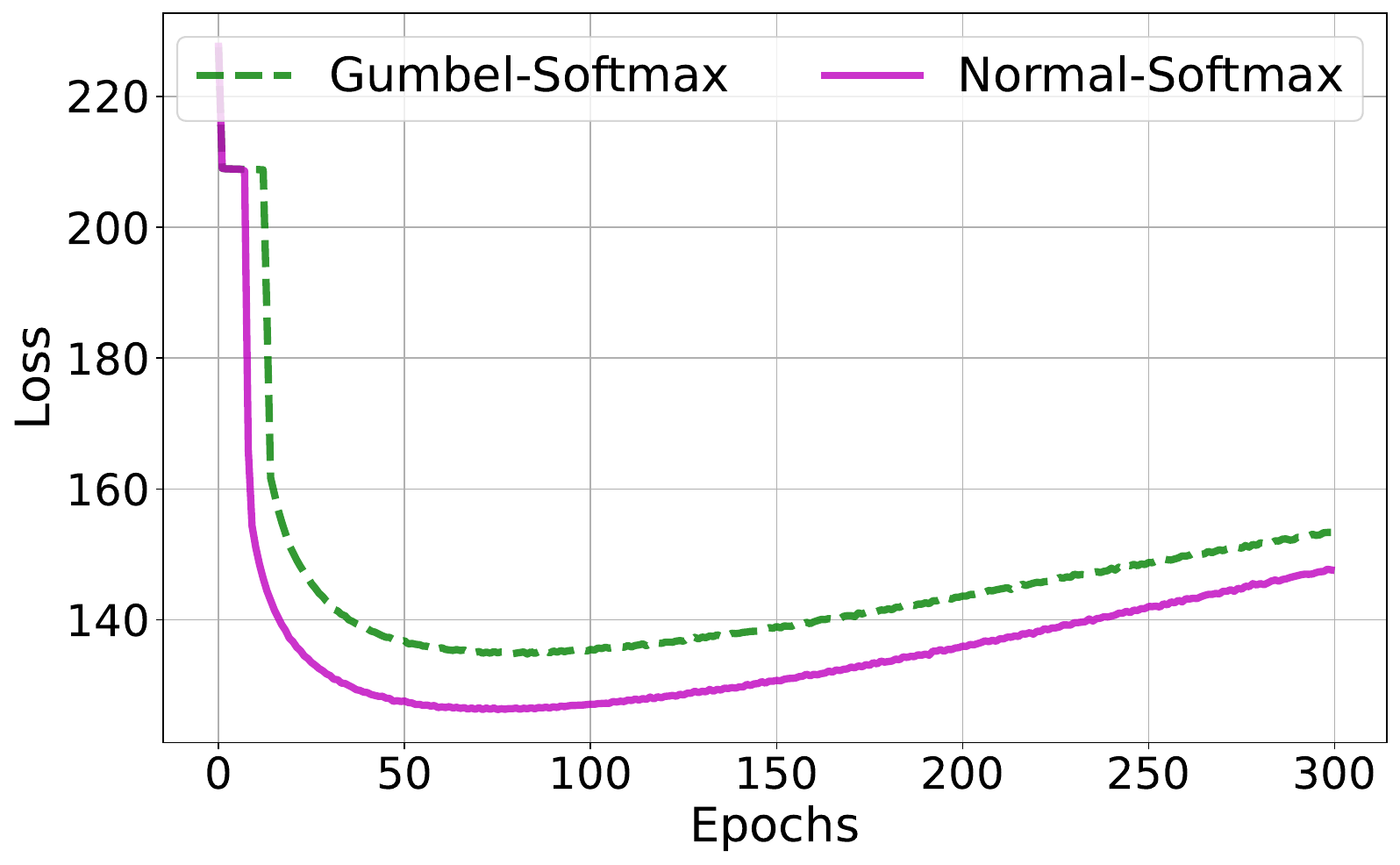}
\end{subfigure}
\begin{subfigure}[b]{.32\textwidth}
\includegraphics[width=\linewidth]{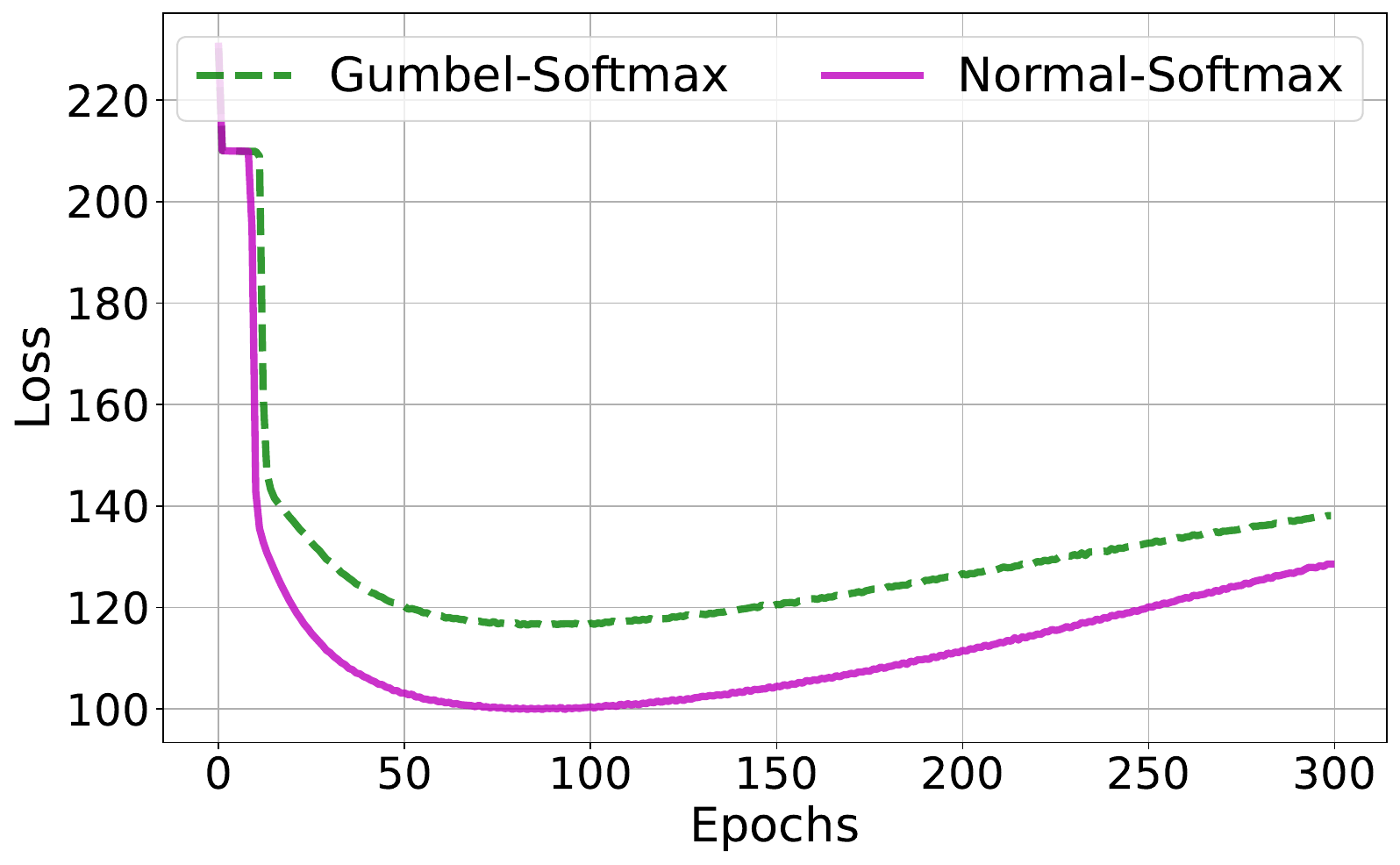}
\end{subfigure}
\begin{subfigure}[b]{.32\textwidth}
\includegraphics[width=\linewidth]{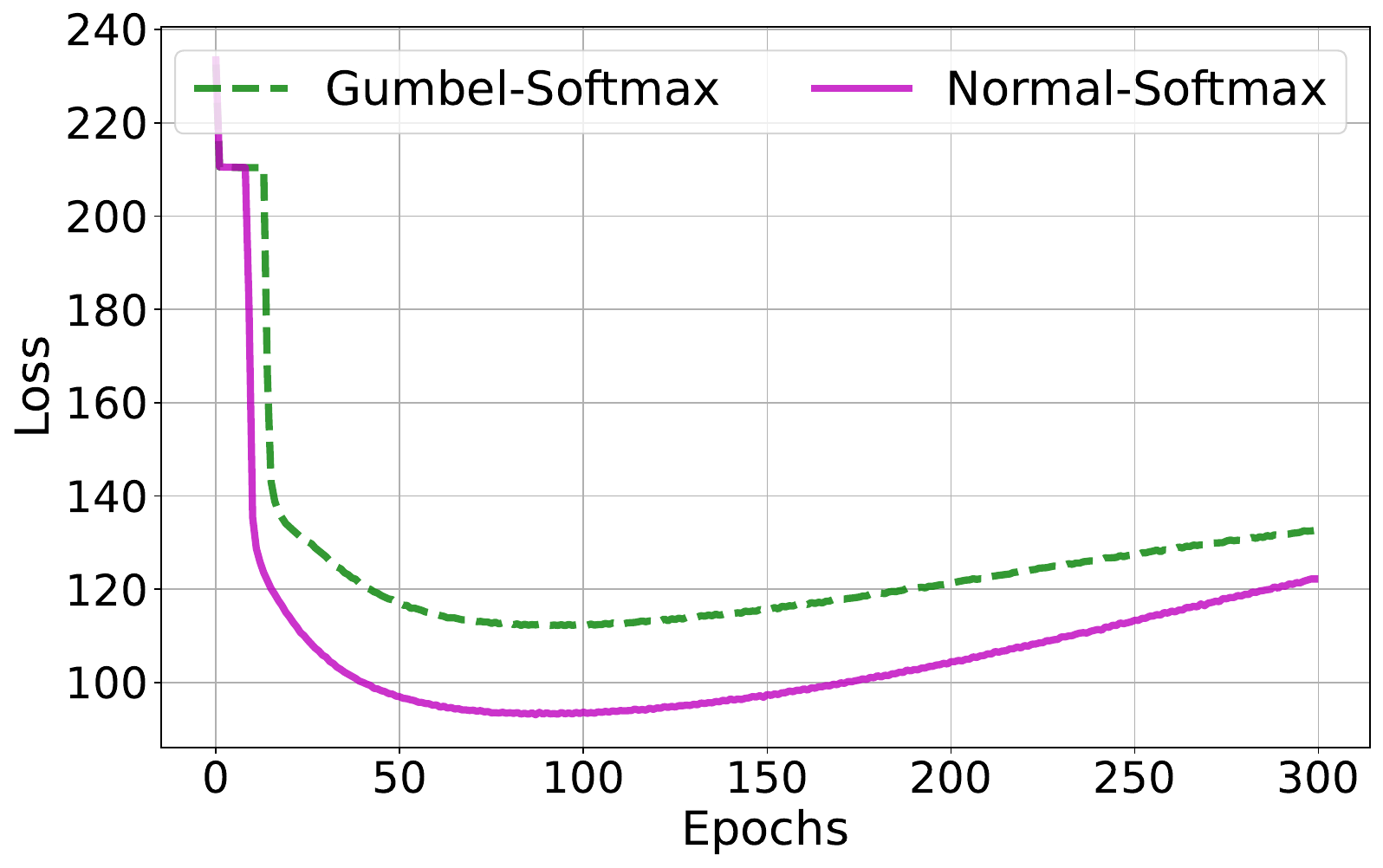}
\end{subfigure}

\begin{subfigure}[b]{.32\textwidth}
\includegraphics[width=\linewidth]{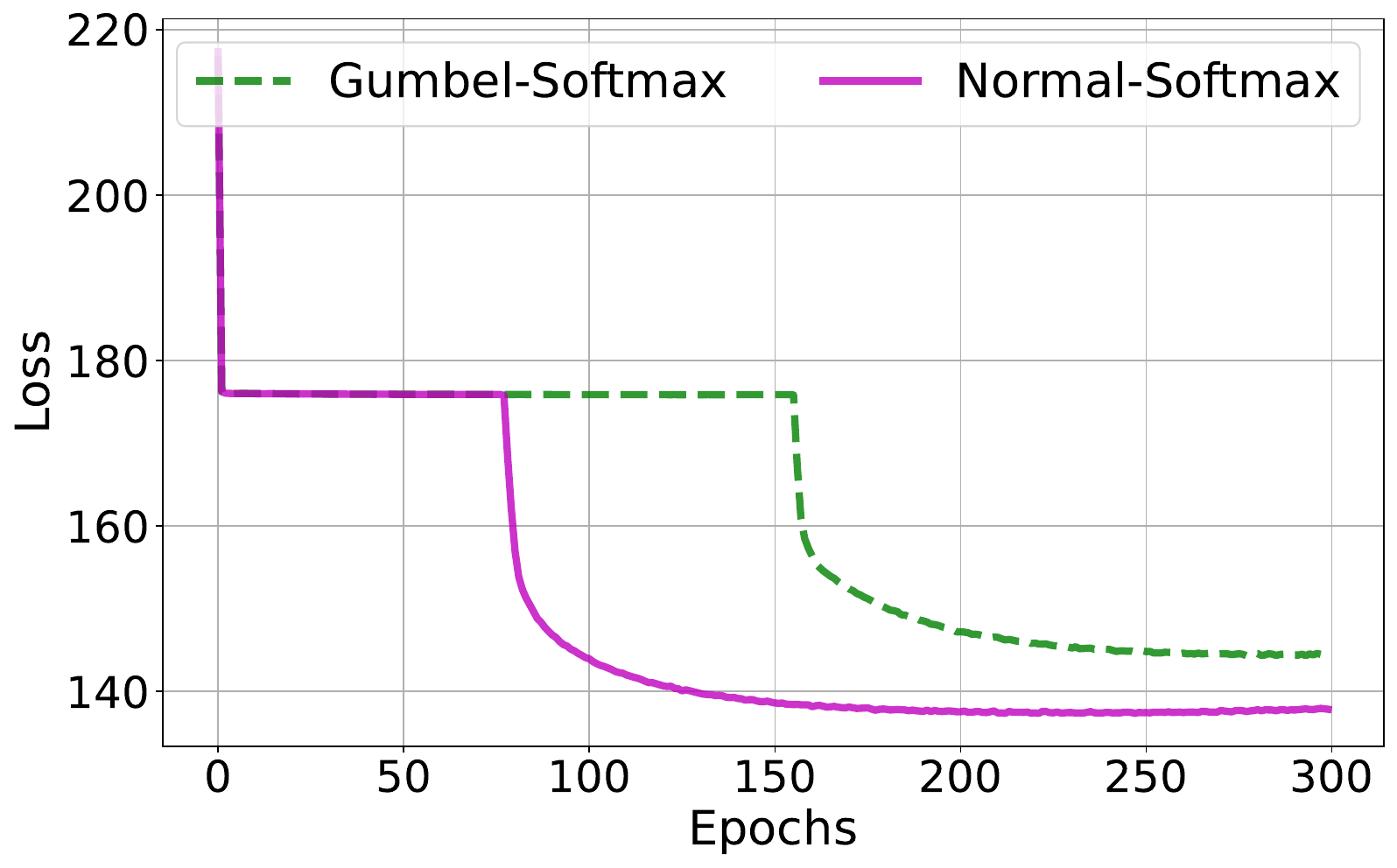}
\caption{K=10}
\end{subfigure}
\begin{subfigure}[b]{.32\textwidth}
\includegraphics[width=\linewidth]{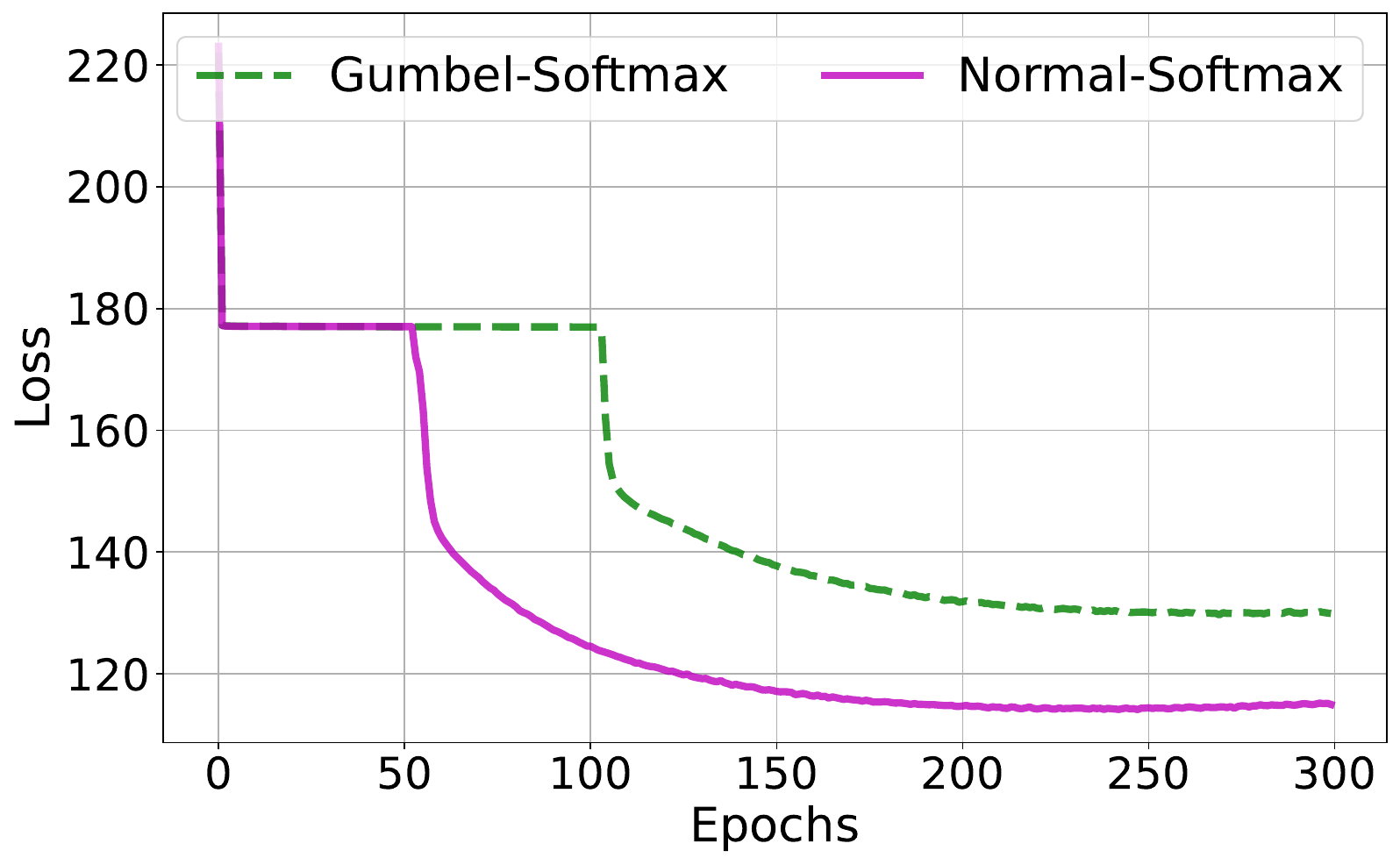}
\caption{K=30}
\end{subfigure}
\begin{subfigure}[b]{.32\textwidth}
\includegraphics[width=\linewidth]{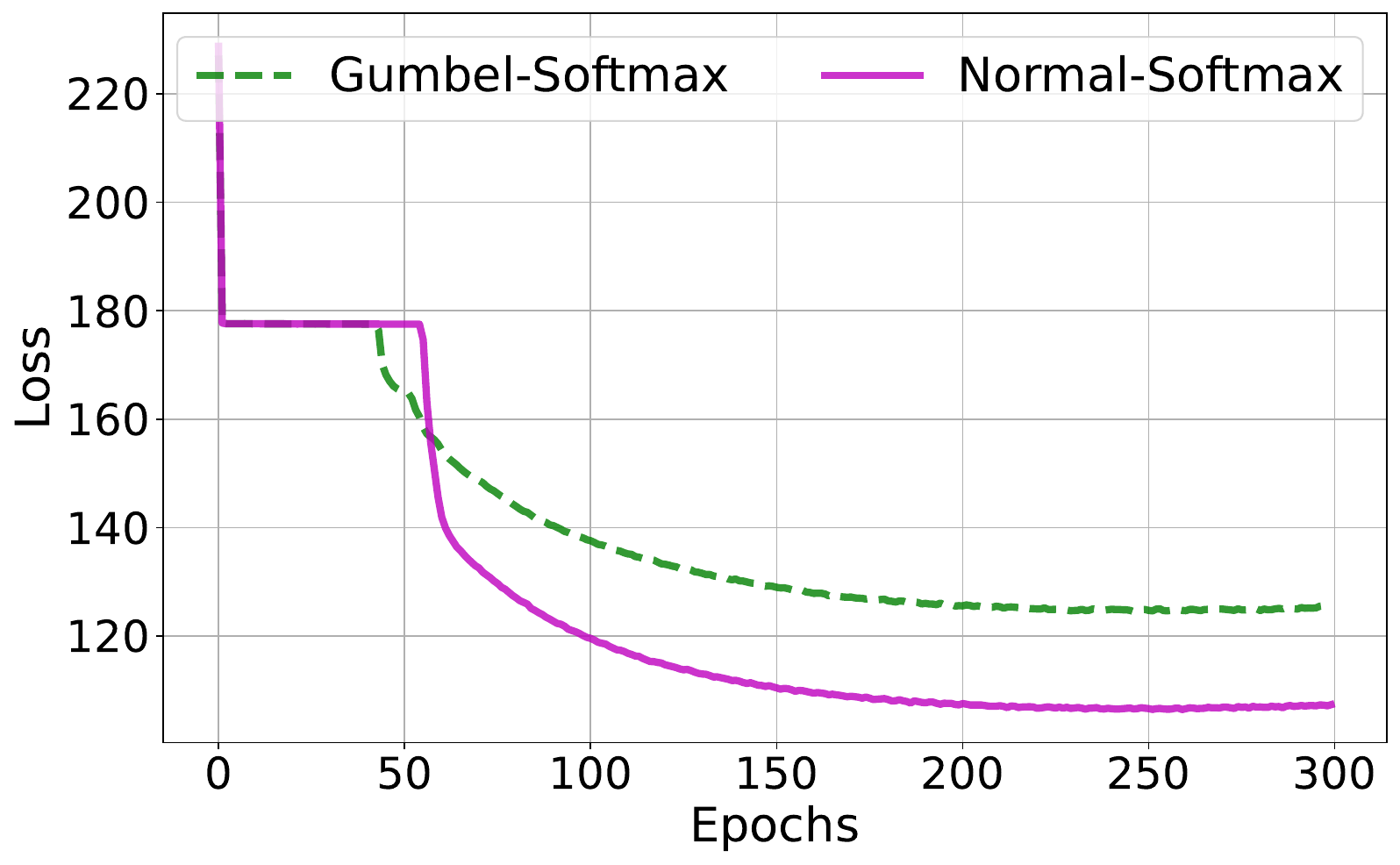}
\caption{K=50}
\end{subfigure}
\caption{Categorical VAE with Perturb-Softmax training loss on the MNIST dataset (top row), and the Omniglot dataset (bottom row) with a $K$-dimensional categorical variable, $K\in [10,30,50]$.}
\label{MNIST-perturb-sm}
\end{figure}

We compared the training ELBO-based loss of categorical Variational-Autoencoders for $N=10 $ variables, each is a $K$-dimensional categorical variable, $K\in [10,30,50]$ on the binarized MNIST \cite{LeCun2005TheMD}, the Fashion-MNIST \citep{xiao2017fashionmnist}, and the Omniglot \citep{doi:10.1126/science.aab3050} datasets for different smooth perturbation distributions.  The architecture consists of an encoder of $X \rightarrow FC(300) \rightarrow \mathrm{ReLU} \rightarrow N*K$, and matching decoder $N*K \rightarrow FC(300) \rightarrow \mathrm{ReLU} \rightarrow  X$. We compare the training convergence when propagating gradients with the Normal-Softmax or the Gumbel-Softmax, both performed with a temperature equal to $1$.
Results show that the former achieves better and faster learning convergence in all experiments (Figures \ref{MNIST-perturb-sm} and  \ref{Fashion-MNIST-perturb-sm}).

\newpage
\bibliographystyle{icml2024}
\bibliography{bibtex}

\onecolumn\section{APPENDIX}
\label{sec:appendix}

\subsection{Related work}
\subsubsection{Convexity}\label{app:related_convexity}
We consider convex functions $f:\Theta \rightarrow \R$ over its convex domain $\Theta = dom(f)$ and follow the notation in \citet{rockafellar-1970a}. A function over a domain $\Theta$ is convex if for every $\theta, \tau \in \Theta$ and $0 \le \lambda \le 1$ it holds that $f(\lambda \theta + (1-\lambda) \tau) \le \lambda f(\theta) + (1-\lambda) f(\tau)$. A function is strictly convex if for every $\theta \ne \tau \in \Theta$ and $0 < \lambda < 1$, it holds that $f(\lambda \theta + (1-\lambda) \tau) < \lambda f(\theta) + (1-\lambda) f(\tau)$. Whenever $f(\theta)$ is twice differentiable, a function is strictly convex if its Hessian is positive definite.  

Convexity is a one-dimensional property. A function $f:\Theta \rightarrow \R$ is convex if and only if its one-dimensional reduction $g(\lambda) \triangleq f(\theta + \lambda v)$ is convex in every admissible direction $v$, i.e., whenever $\theta, \theta + \lambda v \in \Theta$. A twice differentiable function $f(\theta)$ is strictly convex if the second derivative of $g(\lambda)$ is positive in every admissible direction $v$. In this case, we denote $g'(\lambda)$ by $\nabla_v f(\theta)$ and call it a directional derivative:   
\begin{equation}
    \nabla_v f(\theta) \triangleq \lim_{\epsilon \rightarrow 0} \frac{f(\theta + \epsilon v) - f(\theta)}{\epsilon}
\end{equation}
A multivariate function is differentiable if its directional derivative is the same in every direction $v \in \R^d$, namely $\nabla f(\theta) = \nabla_v f(\theta)$ for every $v \in \R^d$.

Convexity admits duality correspondence. Any primal convex function $f(\theta)$ has a dual conjugate function $f^*(p)$
\begin{eqnarray}    
    f^*(p) &=& \max_{\theta \in \Theta} \left\{ \langle p, \theta \rangle - f(\theta) \right\} \label{eq:fstar} \\
    \calP &\triangleq& dom(f^*) \triangleq \{p : f^*(p) < \infty \} \label{eq:domfstar}
\end{eqnarray}
Since $f^*(p)$ is a convex function, its domain $\calP$ is a convex set.

A sub-gradient $p \in \partial f(\theta)$ satisfies $f(\tau) \ge f(\theta) + \langle p, \tau - \theta \rangle$ for every $\tau \in \Theta$. The sub-gradient is intimately connected to directional derivatives. Theorem 23.2 in \citet{rockafellar-1970a} states that 
\begin{equation}\label{rockafellar:23.2}
    p \in \partial f(\theta) \hspace{0.5cm}  \mbox{ iff } \hspace{0.5cm}  \nabla_v f(\theta) \geq \langle p, v \rangle, \;   \forall \mbox{ admissible } v.
\end{equation}
The vector $v$ is admissible if $\theta + \epsilon v \in \Theta$ for small enough $\epsilon$.

The set of all sub-gradients is called sub-differential at $\theta$ and is denoted by $\partial f(\theta)$. A convex function is differentiable at $\theta$ when $\partial f(\theta)$ consists of a single vector, and it is denoted by $\nabla f(\theta)$. 
The sub-differential is a multi-valued mapping between the primal parameters and dual parameters, i.e., 
\begin{equation}
\partial f: \Theta \rightarrow \calP    \label{eq:partialf}
\end{equation}
One can establish with this property the definition of sub-gradient at the optimal point $\theta^* = \arg \max_\theta \{ \langle p, \theta \rangle - f(\theta) \}$. In this case, $0 \in \partial \left( \langle p, \theta^* \rangle - f(\theta^*) \right)$, where the sub-gradient is taken with respect to $\theta$ at the maximal argument $\theta^*$. From the linearity of the sub-gradient, there holds: $p \in \partial f(\theta^*)$, or equivalently, $\{\partial f(\theta) : \theta \in \Theta \} \subseteq \calP $. Using the connection between sub-gradients and directional derivatives, one can show that whenever directional derivatives exist, one can infer a sub-gradient, i.e., the set of all sub-gradients contains the relative interior of $\P$, cf. Theorem 23.4 \citet{rockafellar-1970a}:
\begin{equation}
     ri(\calP) \subseteq \{\partial f(\theta) : \theta \in \Theta \} \subseteq \calP \label{eq:rockaffeller_rangeproof}
\end{equation}

\subsection{Perturb-Softmax probability distributions}

\subsubsection{Completeness and minimality of the Softmax operation}\label{theorem:Gibbs_minimal}
\begin{theorem}
The representation of the softmax distribution defined over $\theta \in \R^d$ is complete. It is minimal when the corresponding log-sum-exp 
\begin{equation}\label{f_theta_logsumexp}
    f(\theta) = \log (\sum_{i=1}^{d} e^{\theta_i})
\end{equation} 
is a strictly convex function (A paraphrase of \citet{WainwrightJordan}, Proposition 3.1).
\end{theorem}
\begin{proof}

First, we note that the derivatives of $f(\theta)$ (Eq. \ref{f_theta_logsumexp}),
\begin{equation}\label{logsumexp_derivative}
    \frac{\partial f(\theta)}{\partial \theta_j} = \softmax(\theta),
\end{equation}
correspond to the softmax probabilities $p^{sm}_\theta$ (Eq. \ref{eq:p_sm}). $p^{sm}_\theta$ is a Gibbs model, hence the representation is complete.  

Let $\theta, \tau \in \R^d$ and denote $e^{\theta_i} = u_i$,  $e^{\tau_i} = v_i$. Then, for $\lambda \in (0,1)$
\begin{eqnarray}
    f(\lambda \theta + (1- \lambda)\tau )&=& \log(\sum_{i=1}^{d} e^{\lambda \theta_i +(1- \lambda)\tau_i }) \\
    &=& \log (\sum_{i=1}^{d}u_i^\lambda v_i^{1-\lambda}) \label{log_sum_eq_dim}.
\end{eqnarray} 
\iffalse From Hölder's inequality, it holds that for $p, q > 1$

\begin{equation}
    \sum_{i=1}^{n} x_iy_i \leq (\sum_{i=1}^{n}|x_i|^p)^{1/p} (\sum_{i=1}^{n}|y_i|^q)^{1/q},
\end{equation}
when $1/p +1/q = 1$. \fi

Applying Hölder's inequality to Equation \ref{log_sum_eq_dim} \iffalse with $1/p= \lambda$ and $1/q = 1-\lambda$  (indeed, for $\lambda \in [0,1]$, it holds that $p \in [1,\infty]$) \fi:
\begin{align}
   \log\left( \sum_{i=1}^{d}u_i^\lambda v_i^{1-\lambda}\right)  &\leq \log(\sum_{i=1}^{d} u_i^{\lambda \frac{1}{\lambda}})^\lambda (\sum_{i=1}^{d}v_i^{(1-\lambda) \frac{1}{1-\lambda}})^{1-\lambda}) \\ &=  \lambda \log(\sum_{i=1}^{d}u_i) + (1-\lambda) \log(\sum_{i=1}^{d} v_i)
   \\ &= \lambda f(\theta)  +(1-\lambda) f(\tau)   . 
\end{align} 
Therefore, it holds that
\begin{equation}\label{eq:convex_f_holder1}
f(\lambda \theta +(1- \lambda)\tau ) \leq \lambda f(\theta)  +(1-\lambda) f(\tau),
\end{equation} proving that $f(\theta)$ is convex.

Hölder's inequality holds with equality if and only if there exists a constant $c\in \R$ such that \iffalse With $x_i = u_i^{\lambda}$, $y_i = v_i^{1-\lambda}$: \fi
\begin{eqnarray}
    |v_i^{1-\lambda}| &=& c |u_i^{\lambda}|^{\frac{1}{\lambda}-1} \quad \forall i \\
    e^{(1-\lambda) \tau_i} &=& c e^{\lambda \theta_i (\frac{1}{\lambda}-1)} \quad \forall i  \\ 
    \iffalse (1-\lambda)\theta'_i  &=& \log(c)+ \lambda\theta_i(\frac{1}{\lambda}-1) \quad \forall i  \\
    (1-\lambda)\tau_i  &=& \log(c)+ (1-\lambda)\theta_i  \quad \forall i \\ \fi
    \tau_i  &=& \frac{\log(c)}{1-\lambda} +\theta_i \quad \forall i 
    \label{eq:convex_f_holder}, 
\end{eqnarray} 
in which case $\theta$ and $\tau$ are linearly constrained and there exists some $\alpha \in \R^2, \alpha \neq 0$ such that $\alpha_1\theta  + \alpha_2 \tau = const$. Therefore, when the representation of $f(\theta)$ is minimal $f(\theta)$ is strictly convex.

Then, consider any $\theta, \tau \in \Theta ^d$, such that  $\Theta = \{ \theta \in \R^d :  \theta_1 = 0\}$. Hölder's inequality holds strictly as there can not exist a constant $c$ such that Equation \ref{eq:convex_f_holder} holds for all $i$ if $d>1$, proving that the representation of $\Theta = \{ \theta \in \R^d:  \theta_1 = 0\}$ is minimal.

Consider $\Theta = \{ \theta \in \R^d :  \sum_j \theta_j = 0\}$. Then, let $\theta, \tau \in \R^d: \sum_i \theta_i = 0, \sum_i \tau_i=0 $, and denote $p_i \propto e^{\theta_i}$ and $q_i \propto e^{\tau_i}$. The proof requires showing that if there exists $i:p_i \neq q_i$ such that $\theta_i  \neq \tau_i+c$. 
Equivalently, it requires proving that if it holds that $\theta_i-\tau_i=0$ for any $i$, then $p_i=q_i$ for all $i$. 
Explicitly, 
\begin{eqnarray}
    p_i&=&\frac{e^\theta_i}{\sum_j e^\theta_j }\\
    q_i&=&\frac{e^\tau_i}{\sum_j e^\tau_j }.
\end{eqnarray}
Then, by marginalization it holds that
\begin{eqnarray}
    \log(p_i)=\log(qi) &\longleftrightarrow & \theta_i-\log(\sum_je^{\theta_j} )=  \tau_i-\log(\sum_je^{\tau_j} )\\
    &\longleftrightarrow & 
    \sum_i \theta_i - \sum_i \log(\sum_je^{\theta_j} ) = \sum_i \tau_i-\sum_i \log(\sum_je^{\tau_j} )\\
    &\longleftrightarrow & d\log(\sum_je^{\theta_j} )= d \log(\sum_je^{\tau_j} ),
\end{eqnarray}
which concludes the proof.

$\Theta = \{ \theta \in \R^d :  \sum_j \theta_j = 0\}$ is complete by the conditions of our completeness theorems by setting $n$ at the $i^{th}$ positions and $-a/d$ everywhere else. 

$\Theta = \{ \theta \in \R^d :  \theta_1=0 \}$ is complete by the conditions of our completeness theorems by setting $n$ at the $1^{st}$ positions and $0$ everywhere else. 
\end{proof}

\begin{corollary}\label{cor:logsumexp_derivative_probability_func}
The derivative of the expected log-sum-exp $f(\theta)$ (Equation \ref{f_theta_logsumexp_perturbed}) is a probability function.
\end{corollary}
\begin{proof}
    Denote $p_{\gamma,j} =  \frac{\partial f(\theta)}{\partial \theta_j}$, then
\begin{eqnarray}\label{log_sum_exp_conjgate_prob}
    \sum_{j=1}^{d} p_{\gamma,j} &=& \sum_{j=1}^{d}\E_\gamma \left[\frac{e^{\theta_j+\gamma_j}}{\sum_{i=1}^{d} e^{\theta_i+\gamma_i}} \right] \\ &=& \E_\gamma \left[\sum_{j=1}^{d}\frac{e^{\theta_j+\gamma_j}}{\sum_{i=1}^{d}  e^{\theta_i+\gamma_i}} \right] 
    = 1.
\end{eqnarray}
Also, 
\begin{equation}\label{log_sum_exp_conjgate_prob_nonneg}
    \E_\gamma \left[\frac{e^{\theta_j+\gamma_j}}{\sum_{i=1}^{d} e^{\theta_i+\gamma_i}} \right] \geq 0  \quad \forall j.
\end{equation}
\end{proof}

\subsection{Supporting proof for Theorem \ref{theorem:p_softmax_complete}}\label{app:completeness_perturbsoftmax}
    First, we prove that $f(\theta)$ (Equation \ref{f_theta_logsumexp_perturbed}) is a closed proper convex function and is also essentially smooth.
    $f(\theta)$ is a convex function as a maximum of convex (linear) functions. Then, 
    $f(\theta)$ is proper as its effective domain is nonempty and it never attains the value $-\infty$, since $\theta \in \R^d$.
     $f(\theta)$ is infinitely differentiable throughout the domain, therefore it is a smooth function throughout its domain. 
     $f(\theta)$ is a smooth convex function on $\R^d$, therefore it is in particular essentially smooth. 
     The smoothness of $f(\theta)$ guarantees its continuity, and since $\R^d$ can be considered a closed set, then $f(\theta)$ is a closed function.

Given the conditions of the theorem on $h_i(\theta)$, we can construct a series $\{ \theta^{(n)} \}_{n=1}^\infty$ for which $h_i(\theta^{(n)}) = n$ for every $n \in \mathbb{N}$. We prove that  $\E_{\gamma} [\arg \max (\theta^{(n)} + \gamma)]$ approaches the zero-one probability vector as $n \rightarrow \infty$. 
\begin{eqnarray}
     \E_\gamma \left[\frac{e^{\theta^{(n)}_i + \gamma_i}}{\sum_{j=1}^d e^{\theta^{(n)}_j + \gamma_j}} \right] &&= \E_\gamma \left[\frac{1}{\sum_{j=1}^d e^{\theta^{(n)}_j + \gamma_j - \theta^{(n)}_i - \gamma_i}} \right] \\
     &&= \E_\gamma \left[ \frac{1}{1 + \sum_{j \ne i} e^{\theta^{(n)}_j - \theta^{(n)}_i + \gamma_j - \gamma_i}} \right] \\
    && \ge \E_\gamma \left[ \frac{1}{1 + \sum_{j \ne i} e^{-n + \gamma_j - \gamma_i}} \right]  \stackrel{n \rightarrow \infty}{\rightarrow 1}
\end{eqnarray}
The limit argument holds since the probability of $\gamma_1,...,\gamma_d$ decay as they tend to infinity. This proves that the zero-one distributions are limit points of probabilities in $\calP$, i.e., $cl(\calP) = \Delta$. 
\subsection{Proof of Lemma \ref{lemma:strictconvexity_softmax}}\label{app:strict_convexity_e_log_sum_exp_proof}
In the following, we prove the strict convexity of the expected log-sum-exp of Lemma \ref{lemma:strictconvexity_softmax}.
\begin{proof}
Let $\theta, \tau \in \R^d$ and $0 < \lambda < 1$. Then 
\begin{eqnarray}
    f(\lambda \theta + (1- \lambda)\tau )= \E_\gamma \left[ \log \left( \sum_{i=1}^{d} e^{\lambda (\theta_i + \gamma_i) +(1- \lambda)(\tau_i + \gamma_i) } \right) \right] 
    = \E_\gamma \left[ \log \left(\sum_{i=1}^{d} u_i v_i \right) \right], \label{eq:perturb_log_sum_eq_dim}
\end{eqnarray} 
where $u_i \triangleq e^{\lambda (\theta_i + \gamma_i)}$ and $v_i \triangleq e^{(1- \lambda)(\tau_i + \gamma_i) }$. Applying H\"{o}lder's inequality $\langle u,v \rangle \le \| v \|_{1/\lambda} \cdot \| u\|_{1/(1-\lambda)}$ we obtain the convexity condition of the log-sum-exp function: 
\begin{equation}
f(\lambda \theta + (1- \lambda)\tau ) \le \lambda f(\theta) + (1-\lambda) f(\tau).
\end{equation}
To prove strict convexity we note that H\"{o}lder's inequality for non-negative vectors $u_i, v_i$ holds with equality if and only if there exists a constant $\alpha \in \R$ such that $v_i = \alpha u_i^{\frac{1 - \lambda}{\lambda}}$ for every $i=1,...,d$, or equivalently:
\begin{equation}
        e^{(1- \lambda)(\tau_i + \gamma_i)} = \left( e^{\lambda (\theta_i + \gamma_i)}\right)^{\frac{1 - \lambda}{\lambda}} \longleftrightarrow \tau_i = \theta_i + c
\end{equation}
Where $c = \frac{\log \alpha}{1-\lambda}$. Therefore, if $\tau \ne \theta + c$, then $\langle u,v \rangle < \| v \|_{1/\lambda} \cdot \| u\|_{1/(1-\lambda)}$ for every $\gamma$ and consequently it also holds when applying the logarithm function and taking an expectation with respect to $\gamma$. 
\end{proof}

\section{Perturb-Argmax probability distributions}

\begin{corollary}\label{cor:derive_probability_max}   
We prove that the derivative of the expected maximizer is the probability of the $\arg \max$. Namely, that $
    \frac{\partial}{\partial \theta}\E_{\gamma  \sim g} \left[ \max_i\{ \theta_i+ \gamma_i\} \right] = P_\gamma \left( \arg \max_i \{ \theta_i+ \gamma_i \} = i \right)$.
\end{corollary}

\begin{proof}
First, by differentiating under the integral:
\begin{equation}
   \partial \E_{\gamma} [ \max \{\theta+\gamma \}] =  \E_{\gamma} [ \partial \max \{\theta+\gamma \}] 
\end{equation}  
Writing a subgradient of the max-function using an indicator function (an application of Danskin’s Theorem):
\begin{equation}\label{eq:argmax_max_derivative}
    \partial \max_i \{\theta_i+ \gamma_i\} = \mathds{1}[\arg \max_i (\theta_i+ \gamma_i) = i] 
\end{equation}  
The proof then follows by applying the expectation to both
sides of Equation \ref{eq:argmax_max_derivative}.
\end{proof}

\subsection{Supporting proof for Theorem \ref{theorem:pert_max_complete}}\label{app:completeness_perturbargmax}
Given the conditions on $h_i(\theta)$, we can construct a series $\{ \theta^{(n)} \}_{n=1}^\infty$ for which $h_i(\theta^{(n)}) = n$ for every $n \in \mathbb{N}$.  We show that $\E_{\gamma} [\arg \max(\theta^{(n)} + \gamma)]$ approaches the zero-one probability vector as $n \rightarrow \infty$. 
\begin{eqnarray}
     \P[i = \arg \max(\theta^{(n)} + \gamma) && = \P \left[ \theta^{(n)}_i + \gamma_i \ge \max_{j \ne i} \{\theta^{(n)}_j + \gamma_j\} \right]  \\
     && \ge  \P \left[\theta^{(n)}_i + \gamma_i \ge \max_{j \ne i} \{\theta^{(n)}_j\} + \max_{j \ne i} \{ \gamma_j \}  \right] \\
    && \ge \P \left[\gamma_i \ge -n + \max_{j \ne i} \{ \gamma_j \} \right] \stackrel{n \rightarrow \infty}{\rightarrow 1} \nonumber
\end{eqnarray}
The limit argument holds since the probability of $\gamma_1,...,\gamma_d$ decay as they tend to infinity.

\begin{corollary}\label{h_max_conjugate}
    The convex conjugate of $f(\theta)$ takes the following values:
    \begin{equation}
    f^{*}\left(\lambda\right) =
        \begin{cases}
            -\E_{\gamma \sim g} \left[ \gamma_{\hat i} \right] &\quad\text{if } \lambda = \lambda^{*}\\
            \infty &\quad\text{otherwise}, 
        \end{cases}
\end{equation}
    where $ \gamma_{\hat i}$ denotes $\gamma_i$ for which $\arg \max _{i} \{ \theta_{i}+ \gamma_{i}  \} = \hat i$.
\end{corollary}

\begin{proof}
    Then, the convex conjugate of $f(\theta)$, when $\lambda^{*}_i $ is denoted by $p_i$ is
\begin{eqnarray}
   f^{*}\left(\lambda\right) &=&  \sum_{i}\theta_i p_i -\E_{\gamma \sim g } \left[ \max_{i} \{ \theta_i+ \gamma_i\} \right] \\
&=&   \sum_{i}\theta_i p_i - \E_{\gamma} \left[ \left(\sum_{\hat i} \mathds{1}_{\arg \max _{i} \{ \theta_{i}+ \gamma_{i}  \} = \hat i}  \right)   \max _{i} \{  \theta_{i}+ \gamma_{i}\} \right] \\
   &=&  \sum_{i}\theta_i p_i - \sum_{\hat i} \E_{\gamma } \left[ \mathds{1}_{\arg \max _{i} \{  \theta_{i}+ \gamma_{i} \} = \hat i} \max _{i} \{ \theta_{i}+ \gamma_{i} \} \right] \\
    &=&  \sum_{i}\theta_i p_i - \sum_{\hat i } \E_{\gamma } \left[ \mathds{1}_{\arg \max _{i} \{ \theta_{i}+ \gamma_{i} \} = \hat i}  (\theta_{\hat i}+ \gamma_{\hat i}) \right] \\
    &=&  \sum_{i}\theta_i p_i - \sum_{\hat i} \E_{\gamma} \left[ \mathds{1}_{\arg \max _{i} \{\theta_{i}+ \gamma_{i} \} = \hat i}  \theta_{\hat i} \right] -\sum_{\hat i} \E_{\gamma} \left[ \mathds{1}_{\arg \max _{i} \{ \theta_{i}+ \gamma_{i} \} = \hat i} \gamma_{\hat i} \right]  \\
    &=&  \sum_{i}\theta_i p_i - \sum_{\hat i } p_{\hat i} \theta_{\hat i}  -\sum_{\hat i} \E_{\gamma} \left[ \mathds{1}_{\arg \max _{i} \{ \theta_{i}+ \gamma_{i}  \} = \hat i}  \gamma_{\hat i} \right] \\
     &=&   -\sum_{\hat i} \E_{\gamma} \left[ \mathds{1}_{\arg \max _{i} \{ \theta_{i}+ \gamma_{i}  \} = \hat i}  \gamma_{\hat i} \right] \\
     &=&  -\sum_{\hat i} p_{\hat i} \E_{\gamma} \left[ \gamma_{\hat i} \right] \\
      &=&   -\E_{\gamma} \left[ \gamma_{\hat i} \right] 
\end{eqnarray}

where $ \gamma_{\hat i}$ denotes $\gamma_i$ for which $\arg \max _{i} \{ \theta_{i}+ \gamma_{i}  \} = \hat i$.

\end{proof}

\subsection{Proof of Lemma \ref{lemma:diffofPerturb-Max}}\label{app:diffofPerturb-Max}

\begin{proof}
By reparameterization
\begin{eqnarray}
    f(\theta) &=& \int_{\R^d} p(\gamma) \max \{ \theta + \gamma\} d\gamma
    = \int_{\R^d} p(\theta - \gamma) \max \{ \gamma\} d\gamma
    \end{eqnarray}
The proof concludes by differentiating under the integral sign while noting that $p(\theta-\gamma)$ is differentiable. 
\end{proof}

\subsection{Proof of Theorem \ref{theorem:p_argmax_minimal}}\label{app:minnimality_perturb_argmax}
In what follows we prove the minimality of the Perturb-Argmax of Theorem \ref{theorem:p_argmax_minimal}.
\begin{proof}
    Similarly to the Perturb-Softmax setting, we prove that under these conditions the function $f(\theta) = \E_\gamma [\max \{\theta + \gamma\}]$ is strictly convex. For this we rely on its one dimensional function $g(\lambda) \triangleq f(\theta + \lambda v)$ and show that $g''(\lambda) > 0$. Since the function $g(\lambda)$ is convex then $g''(\lambda) \ge 0$, and it is enough to show that $g'(\lambda)$ depends on $\lambda$, for which it follows that $g''(\lambda) \ne 0$ and consequently $g''(\lambda) > 0$. 

    $g'(\lambda)$ is the directional derivative $\nabla_v f(\theta)$ in every admissible direction $v = \tau - \theta$, for $\tau, \theta \in \Theta$. The theorem conditions assert that $v$ is not the constant vector, i.e., $v \ne c1$, where $c$ is some constant and $1 = (1,...,1)$ is the all-one vector. 

    We assume, without loss of generality, that $\max\{ \theta + \gamma\} \triangleq \max_{i=1,...,d} \{ \theta_i + \gamma_i \}$ is chosen between two indexes, namely $\max\{ \theta + \gamma \} = \max\{ \theta_1 + \gamma_1, \theta_j + \gamma_j \}$. This is possible as we treat $j$ to be $j = \arg \max_{i \ne 1} \{\theta_j + \gamma_j\}$. We denote by $p_1(\gamma_1)$ the differentiable probability density function of $\gamma_1$. We denote remaining random variables as 
    $\gamma_{-1} \triangleq (\gamma_2,...,\gamma_d)$, their probability density function by their measure $d\mu(\gamma_{-1})$. 

    We analyze $g'(\lambda) = \lim_{\epsilon \rightarrow 0} \frac{1}{\epsilon} (g(\lambda + \epsilon) - g(\lambda))$
\begin{eqnarray}
    g(\lambda) &=& E_\gamma [ \max\{\theta_1 + \lambda v_1 + \gamma_1  , \theta_j + \lambda v_j + \gamma_j \}]  \\
    &=& \int d\mu(\gamma_{-1}) \int_{-\infty}^{\alpha(\gamma_j)} p_1(\gamma_1) (\theta_j + \lambda v_j + \gamma_j) d\gamma_1 + \int d\mu(\gamma_{-1}) \int_{\alpha(\gamma_j)}^\infty p_1(\gamma_1) (\theta_1 + \lambda v_1 + \gamma_1) d\gamma_1 \nonumber
\end{eqnarray}
$\alpha(\gamma_j)$ is the threshold for which $\gamma_1$ shifts the maximal value to $\theta_1 + \lambda v_1 + \gamma_1$, namely $ \alpha(\gamma_j) = \theta_j + \lambda v_j + \gamma_j - \theta_1 - \lambda v_1$. 

With this notation, $g(\lambda + \epsilon)=$   \begin{equation}
     \int d\mu(\gamma_{-1}) \int_{-\infty}^{\alpha(\gamma_j) + \epsilon(v_j - v_1)}  \hspace{-1.7cm}  p_1(\gamma_1) (\theta_j + (\lambda + \epsilon) v_j + \gamma_j) d\gamma_1 
      + \int d\mu(\gamma_{-1}) \int_{\alpha(\gamma_j) + \epsilon (v_j-v_1)}^\infty  \hspace{-1.7cm}  p_1(\gamma_1) (\theta_1 + (\lambda +\epsilon) v_1  + \gamma_1) d\gamma_1  
\end{equation}

Their difference is composed of four terms: 
\begingroup
\allowdisplaybreaks
\begin{align}
& g(\lambda+\epsilon) - g(\lambda) = \nonumber \\
    &  \int d\mu(\gamma_{-1}) \int_{-\infty}^{\alpha(\gamma_j) + \epsilon(v_j - v_1)}  \hspace{-1.7cm}  p_1(\gamma_1) \epsilon v_j d\gamma_1 
     + \int d\mu(\gamma_{-1}) \int_{\alpha(\gamma_j) + \epsilon (v_j-v_1)}^\infty  \hspace{-1.7cm}  p_1(\gamma_1) \epsilon v_1  d\gamma_1 \nonumber \\   
    &+ \int d\mu(\gamma_{-1}) \int_{\alpha(\gamma_j)}^{\alpha(\gamma_j) + \epsilon(v_j - v_1)}  \hspace{-1.7cm}  p_1(\gamma_1) (\theta_j + (\lambda + \epsilon) v_j + \gamma_j) d\gamma_1 \nonumber \\
    &   - \int d\mu(\gamma_{-1}) \int_{\alpha(\gamma_j)}^{\alpha(\gamma_j) + \epsilon (v_j-v_1)}  \hspace{-1.7cm}  p_1(\gamma_1) (\theta_1 + (\lambda +\epsilon) v_1  + \gamma_1) d\gamma_1 \nonumber    
    \end{align}
\endgroup    
Taking the limit to zero $\lim_{\epsilon \rightarrow 0} \frac{1}{\epsilon} (g(\lambda + \epsilon) - g(\lambda)$, the last two terms cancel out, since when taking the limit then $\gamma_1 = \alpha(\gamma_j)$ and $ \alpha(\gamma_j) = \theta_j + \lambda v_j + \gamma_j - \theta_1 - \lambda v_1$ by definition, or equivalently, $\alpha(\gamma_j) +  \theta_1 + \lambda v_1 = \theta_j + \lambda v_j + \gamma_j $. Therefore 
\begin{align}
    g'(\lambda) = \int d\mu(\gamma_{-1}) \int_{-\infty}^{\alpha(\gamma_j)} p_1(\gamma_1)  v_j d\gamma_1
    + \int d\mu(\gamma_{-1}) \int_{\alpha(\gamma_j)}^\infty p_1(\gamma_1) v_1  d\gamma_1
\end{align}
We conclude that by the conditions of the theorem $\alpha(\gamma_j) = \theta_j + \lambda v_j + \gamma_j - \theta_1 - \lambda v_1$ is a function of $\lambda$ since there exists $j$ for which $v_j - v_1 \ne 0$ and the probability density function $p_1(\gamma_1) > 0$ therefore it assigns mass on the intervals $[-\infty, \alpha(\gamma_j)]$ and $[\alpha(\gamma_j), \infty]$. Therefore $g'(\lambda)$ is non-constant function of $\lambda$ and $g''(\lambda) \ne 0$. 
\end{proof}
\subsection{Proof of Proposition \ref{prop:max_discrete}} \label{app:h_max_discrete}

Let $\theta \in \R^2$, and consider i.i.d. random variables $\gamma \in \{1,-1\}$, such that $P(\gamma = 1)=P(\gamma=-1)=\frac{1}{2}$. Let $f(\theta) = \E_{\gamma } \left[ \max_{i}\{ \theta_i+ \gamma_i\} \right]$ denote the expected perturbed maximum over the domain $\R^d$. Let $f(\theta)$ take values over the extended real domain $\R \cup \{\pm \infty\}$. Clearly, $P(1= \arg \max_i \{\theta_i+\gamma_i \}) = P(\theta_1+\gamma_1 \geq \theta_2+\gamma_2)$.

Then, $f(\theta)$ can be explicitly expressed as:
\begin{eqnarray}
    f(\theta) &=& \E_{\gamma } \left[ \max_i\{ \theta_i+ \gamma_i\} \right] \\
    &=& P(\gamma_1=1,\gamma_2 =1)\left(1+ \max\{\theta_1,\theta_2\}\right) + P(\gamma_1=-1,\gamma_2 =-1)\left(\max\{\theta_1, \theta_2\}-1\right)  \nonumber\\
    &+& P(\gamma_1=1,\gamma_2 =-1)\left(
    \max\{\theta_1+1, \theta_2-1\}\right) +P(\gamma_1=-1,\gamma_2 =1)\left(\max\{\theta_1-1, \theta_2+1\}\right) \nonumber \\   
    &=& \frac{1}{2}\left(\max\{\theta_1, \theta_2\}\right)
    + P(\gamma_1=1,\gamma_2 =-1)\left(
    \max\{\theta_1+1, \theta_2-1\}\right) +P(\gamma_1=-1,\gamma_2 =1)\left(\max\{\theta_1-1, \theta_2+1\}\right) \nonumber \\
    &=& \frac{1}{2}\left(\max\{\theta_1, \theta_2\}\right)
    + \frac{1}{2}\left(\frac{\max\{\theta_1+1, \theta_2-1\}}{2}+\frac{\max\{\theta_1-1, \theta_2+1\}}{2}\right) \nonumber \label{max_discrete_f}
\end{eqnarray}

Equation \ref{max_discrete_f} suggests that $f(\theta)$ takes the following form:

\begin{equation}\label{discrete_max_ranges}
    f(\theta) =
        \begin{cases}
        \theta_1 &\quad\text{if } \theta_1 \geq \theta_2+2\\

        \frac{3}{4}\theta_1  + \frac{1}{4}\theta_2+ \frac{1}{2} &\quad\text{if }  \theta_2+2 \geq \theta_1 \geq \theta_2\\

        \frac{3}{4}\theta_2  + \frac{1}{4}\theta_1+ \frac{1}{2} &\quad\text{if }\theta_2-2 \leq  \theta_1 \leq  \theta_2\\
        
        \theta_2 &\quad\text{if } \theta_1 \leq \theta_2-2
        \end{cases}
\end{equation}

\begin{figure}[t]
\centering
\includegraphics[width=0.28\textwidth]{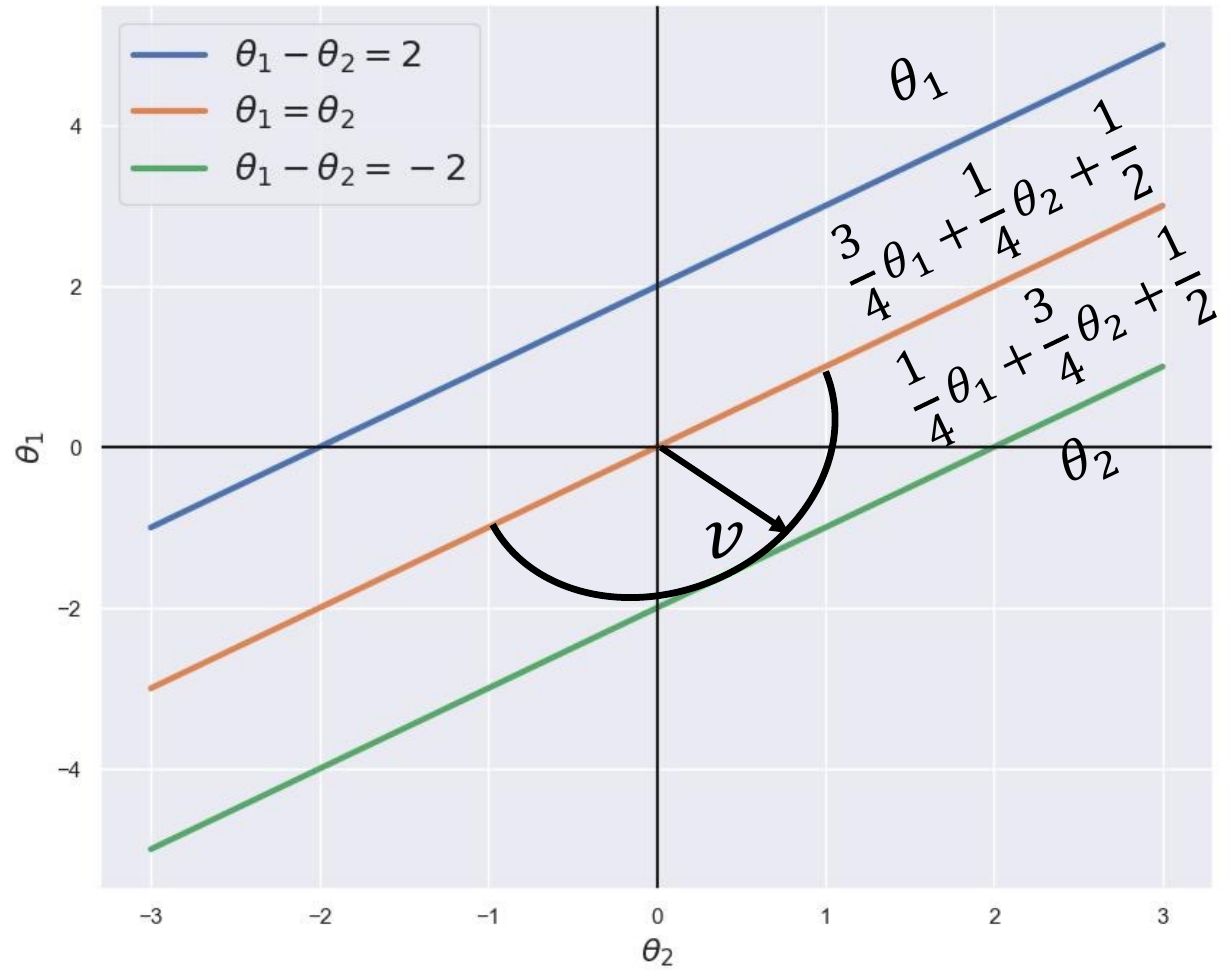}\caption{An illustration of the function $f(\theta)$ (Equation \ref{discrete_max_ranges}) and the direction vector $v: v_1 \geq v_2$ at $\theta=(0,0)$.}
\label{app:subgradient_illustr}
\end{figure}
Recall, that we aim to prove that the Perturb-Argmax probability model is unidentifiable. The function $f(\theta)$, illustrated in Figure \ref{app:subgradient_illustr} in the appendix, is continuous and differentiable almost everywhere. 
However, in its overlapping segments, i.e., when $\theta_1 = \theta_2 + 2$, $\theta_1 = \theta_2$ and $\theta_1 + 2 = \theta_2$, the function is not differentiable, i.e., it has a sub-differential $\partial f(\theta)$ which is a set of sub-gradients. To prove that the Perturb-Argmax probability model is unidentifiable, we show that $\partial f(\theta)$ is a multi-valued mapping when $\theta_1 = \theta_2$. In particular, we show that every probability distribution $p = (p_1,p_2)$ with $p_1 \in [\frac{1}{4}, \frac{3}{4}]$ satisfies $p \in \partial f(\theta)$. 

For this task, we recall the connection between sub-gradients and directional derivatives: $p \in \partial f(\theta)$ if $\nabla_v f(\theta) \ge \langle p, v \rangle$ for every $v \in \R^2$. 
When $\theta_1 = \theta_2 = c$, then $f(\theta) = c + \frac{1}{2}$, thus for the direction $v = (v_1, v_2)$ for which $v_1 \ge v_2$ holds $\nabla_v f(\theta) = \frac{3}{4} v_1 + \frac{1}{4} v_2$. 
Recall that $p \in \partial f(\theta)$ if $ \frac{3}{4} v_1 + \frac{1}{4} v_2 \ge \langle p, v \rangle$ for every $v_1 \ge v_2$. Thus we conclude that $p$ must satisfy $p_1 \le \frac{3}{4}$. Since the same holds to $v_2 \ge v_1$ then $p_1 \ge \frac{1}{4}$. Taking both these conditions, $p \in \partial f(\theta)$ when $\frac{1}{4} \le p_1 \le \frac{3}{4}$. 
Therefore, $\partial f(\theta)$ is multi-valued mapping, or equivalently, the parameters $\theta = (\theta_1, \theta_2)$ are not identifying probability distributions. The sub-differential mapping is 

\begin{equation}\label{discrete_argmax_ranges_prob}
    (\partial f(\theta))_1 =
        \begin{cases}
        1 &\text{if } \theta_1 >  \theta_2+2\\
        [\frac{3}{4},1] &\text{if } \theta_1 = \theta_2+2\\
        \frac{3}{4} &\text{if }  \theta_2+2 >   \theta_1 >  \theta_2\\
        [\frac{1}{4},\frac{3}{4}] &\text{if } \theta_1 = \theta_2\\
        \frac{1}{4} &\text{if }\theta_2-2 < \theta_1 < \theta_2\\
        [0,\frac{1}{4}] &\text{if } \theta_1 = \theta_2 -2\\
        0  &\text{if } \theta_1 < \theta_2-2.
        \end{cases},
\end{equation}
as illustrated in Figure \ref{fig:map_prob_illustr}.

\subsection{Proof of Proposition \ref{prop:nonminimal_argmax_bounded}}\label{app:max_sbounded}

The perturb-max function $f(\theta)$ can be expressed as
\begin{align}
    f(\theta) &= \E_{\gamma_1,\gamma_2 \sim [-1,1]} \left[ \max_i\{ \theta_i+ \gamma_i\} \right] \\
    &= \E_{\gamma_1,\gamma_2} \left[ \max\{ \theta_1+ \gamma_1 ,\theta_2+ \gamma_2  \} \right]  - E_{\gamma_2 } \left[ \gamma_2 \right]\\
    &= \E_{\gamma_1,\gamma_2 } \left[ \max\{ \theta_1+ \gamma_1 ,\theta_2+ \gamma_2  \}  - \gamma_2 \right] \\
    &= \E_{\gamma_1,\gamma_2 } \left[ \max\{ \theta_1+ \gamma_1 -\gamma_2 ,\theta_2 \}  +\theta_2-\theta_2 \right] \\
    &=\theta_2 + \E_{\gamma_1,\gamma_2} \left[ \max\{ \theta_1-\theta_2 + \gamma_1 -\gamma_2 ,0 \}  \right], 
\end{align}
when the second equation holds since $E_{\gamma_2 \sim [-1,1]} \left[ \gamma_2 \right] =0$, and the distribution of $\gamma$ is omitted for brevity.

Define $\theta = \theta_1-\theta_2$ and $Z = \gamma_1-\gamma_2$. Then, the random variable $Z$ has a triangular distribution. The random variable $Z$ has the following cdf:
\begin{align}
    F_{Z}(z) &= P(\gamma_1-\gamma_2 \leq z) \\
    &=
    \begin{cases}
        0 &\quad\text{if }  z < -2 \\ 
        \frac{1}{4}\frac{(2+z)^2}{2} &\quad\text{if } 0 > z \geq -2\\
         \frac{1}{4}(4-\frac{(2-z)^2}{2}) &\quad\text{if } 2 \geq z \geq 0 \\ 
        0 &\quad\text{if }  z > 2 
    \end{cases}
\end{align}

\begin{figure}[h]
\centering
\centering\includegraphics[width=0.25\linewidth]{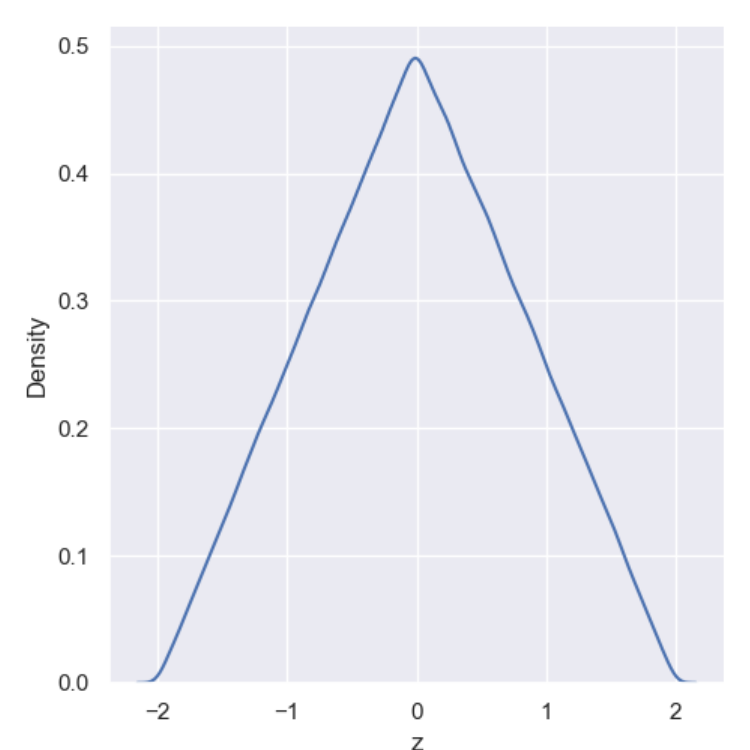}
\caption{Simulation of the density of the difference between 1M iid uniform(-1,1) random variables.}
\label{fig:simulated_pdf_z}
\end{figure}

The random variable $Z$ has the following pdf, also verified in simulation of the density of the difference between 1M iid $U(-1,1)$ random variables (Figure \ref{fig:simulated_pdf_z}):
\begin{eqnarray}\label{pdf_smoothbounded}
    f_{Z}(z) &=& 
    \begin{cases}
        \frac{1}{4}(2+z) \quad\text{if } 0 > z \geq -2\\
        \frac{1}{4}(2-z) \quad\text{if } 2 \geq z \geq 0 \\ 
         0 \quad\text{otherwise } 
    \end{cases}
\end{eqnarray}

With the pdf of the random variable $Z$ (Equation \ref{pdf_smoothbounded}) consider $f(\theta)$ the appropriate range of $\theta$.

\begin{enumerate}
\item \textbf{Case: $-\theta < -2$} 

In this case $\theta_1-\theta_2 > 2$, therefore
\begin{align}
    f(\theta) &= \E_{\gamma_1,\gamma_2 \sim [-1,1]} \left[ \max_i\{ \theta_i+ \gamma_i\} \right] \\
    &= \E_{\gamma_1 \sim [-1,1]} \left[ \theta_1+ \gamma_1\ \right] \\ 
    &= \theta_1+ \E_{\gamma_1 \sim [-1,1]} \left[\gamma_1\ \right] \\
    &= \theta_1
\end{align}

\item \textbf{Case: $0 \leq -\theta \leq 2$} 

{\allowdisplaybreaks
\begin{align}
    f(\theta) &= \theta_2 + \E_{z} \left[ \max\{ \theta + z ,0 \}  \right] \\
    &= \theta_2 + \int_{-\theta}^{2} f_Z(z)( \theta + z) dz \label{eq:f_smoothbounded1}\\
    &= \theta_2 + \int_{-\theta}^{2} \frac{1}{4}(2-z)( \theta + z) dz \\
    %&= \theta_2 + \int_{-\theta}^{2} \frac{1}{4}(2\theta +2z - z\theta -z^2) dz \\
    &= \theta_2 + \frac{1}{2} \int_{-\theta}^{2} \theta dz + \frac{1}{2} \int_{-\theta}^{2}  z dz - \frac{1}{4}\int_{-\theta}^{2} z \theta dz - \frac{1}{4}\int_{-\theta}^{2}  z^2 dz  \\
    %&= \theta_2 +  \frac{1}{2}\left[  \theta z \right]_{-\theta}^2 + \frac{1}{2} \left[  \frac{z^2}{2} \right]_{-\theta}^2 - \frac{1}{4} \left[  \frac{z^2\theta}{2} \right]_{-\theta}^2 - \frac{1}{4} \left[  \frac{z^3}{3} \right]_{-\theta}^2 \\
    %&= \theta_2 + \frac{1}{2}(2+\theta)\theta +1- \frac{\theta^2}{4} -(\frac{1}{2}\theta- \frac{\theta^3}{8}) - (\frac{2}{3}+\frac{\theta^3}{12})\\
    %&= \theta_2 + \frac{1}{3} + \frac{1}{2}\theta +\frac{1}{4}\theta^2 + \frac{1}{24}\theta^3 \\
    %&= \theta_2 +  \frac{1}{4} \left(\frac{4}{3} +2\theta + \theta^2 +\frac{\theta^3}{6} \right)\\
   %&= \theta_2 +  \frac{1}{4} \left(\frac{4}{3} +2(\theta_1-\theta_2) + (\theta_1-\theta_2)^2 +\frac{(\theta_1-\theta_2)^3}{6} \right)\\
    %&= \theta_2 +  \frac{1}{4} \left(\frac{4}{3} +2\theta_1-2\theta_2+ \theta_1^2+\theta_2^2-2 \theta_1 \theta_2 + \frac{\theta_1^3 - 3 \theta_1^2 \theta_2+ 3 \theta_1 \theta_2^2 - \theta_2^3}{6} \right) \\
    %&= \frac{1}{4} \left(\frac{4}{3} +2\theta_1 + 2\theta_2+ \theta_1^2+\theta_2^2-2 \theta_1 \theta_2 + \frac{\theta_1^3 - 3 \theta_1^2 \theta_2+ 3 \theta_1 \theta_2^2 - \theta_2^3}{6} \right) \\
     &= \frac{1}{4} \left(\frac{4}{3} +2(\theta_1 + \theta_2)+ (\theta_1-\theta_2)^2 + \frac{1}{6}(\theta_1-\theta_2)^3\right)
\end{align}
}

\item \textbf{Case: $-2 \leq -\theta \leq 0$}  
{\allowdisplaybreaks
\begin{align}
    f(\theta) &= \theta_2 + \E_{z} \left[ \max\{ \theta + z ,0 \}  \right] \\
    &= \theta_2 + \int_{-\theta}^{0} f_Z(z)( \theta + z) dz + \int_{0}^{2} f_Z(z)( \theta + z) dz \\
    &= \theta_2 + \int_{-\theta}^{0} \frac{1}{4}(2+ z)(\theta +z) dz + \int_{0}^{2} \frac{1}{4}(2- z)(\theta +z) dz  \\
     &= \theta_2 + \frac{1}{4}(\theta^2- \frac{1}{6} \theta^3) + \frac{1}{4}(2 \theta +\frac{4}{3} ) \\
     %&= \theta_2 + \frac{1}{4}(\theta^2- \frac{1}{6} \theta^3 + 2 \theta +\frac{4}{3} )\\
     &= \frac{1}{4} \left(\frac{4}{3}+ 2 (\theta_1+\theta_2) +(\theta_1-\theta_2)^2 -\frac{1}{6}(\theta_1-\theta_2)^3 \right)
\end{align}

\begin{align}
   \int_{-\theta}^{0}  \frac{1}{4}(2+ z)(\theta +z)dz  &= \frac{1}{2} \int_{-\theta}^{0} \theta dz + \frac{1}{2} \int_{-\theta}^{0}  z dz + \frac{1}{4}\int_{-\theta}^{0} z \theta dz + \frac{1}{4}\int_{-\theta}^{0} z^2 dz   \\
    &= \frac{1}{2}\left[  \theta z \right]_{-\theta}^0 + \frac{1}{2} \left[  \frac{z^2}{2} \right]_{-\theta}^0 + \frac{1}{4} \left[  \frac{z^2\theta}{2} \right]_{-\theta}^0 + \frac{1}{4} \left[  \frac{z^3}{3} \right]_{-\theta}^0 \\ 
    %&= \frac{1}{2} \theta^2 - \frac{1}{4}\theta^2 -\frac{1}{8} \theta^3 +\frac{1}{12}\theta^3 \\
    &= \frac{1}{4}(\theta^2- \frac{1}{6} \theta^3)
\end{align}

\begin{align}
    \int_{0}^{2} \frac{1}{4}(2-z)( \theta + z) dz
    &=  \int_{0}^{2}  \frac{1}{4}(2\theta +2z - z\theta -z^2) dz \\
    &=  \frac{1}{2} \int_{0}^{2} \theta dz + \frac{1}{2} \int_{0}^{2} z dz - \frac{1}{4}\int_{0}^{2} z \theta dz - \frac{1}{4}\int_{0}^{2}  z^2 dz  \\
    &=  \frac{1}{2}\left[  \theta z \right]_0^2 + \frac{1}{2} \left[  \frac{z^2}{2} \right]_0^2 - \frac{1}{4} \left[  \frac{z^2\theta}{2} \right]_0^2- \frac{1}{4} \left[  \frac{z^3}{3} \right]_0^2 \\
    %&=  \theta +1 - \frac{1}{2} \theta -\frac{2}{3} \\
    %&= \frac{1}{2} \theta +\frac{1}{3} \\
    &= \frac{1}{4}(2 \theta +\frac{4}{3} )
\end{align}
    
\item \textbf{Case: $-\theta > 2$} 

In this case $\theta_1-\theta_2 < -2$,  therefore
\begin{align}
    f(\theta) &= \E_{\gamma_1,\gamma_2 \sim [-1,1]} \left[ \max_i\{ \theta_i+ \gamma_i\} \right] \\
    &= \E_{\gamma_2 \sim [-1,1]} \left[ \theta_2+ \gamma_2\ \right] \\ 
    &= \theta_2+ \E_{\gamma_2 \sim [-1,1]} \left[\gamma_2\ \right] \\
    &= \theta_2
\end{align}
}
\end{enumerate}

To conclude, 
\begin{align}\label{ftheta_boundedsmooth}
    f(\theta) &=
    \begin{cases}
        \theta_1 \quad\text{if }  -\theta < -2 \\ 
        \frac{1}{4} \left(\frac{4}{3} +2(\theta_1 + \theta_2)+ (\theta_1-\theta_2)^2 + \frac{1}{6}(\theta_1-\theta_2)^3\right)    \quad\text{if } 0 \leq -\theta \leq 2 \\
        \frac{1}{4} \left(\frac{4}{3}+ 2 (\theta_1+\theta_2) +(\theta_1-\theta_2)^2 -\frac{1}{6}(\theta_1-\theta_2)^3 \right) \quad\text{if } -2 \leq -\theta \leq 0 \\ 
        \theta_2 \quad\text{if }  -\theta > 2
    \end{cases}
\end{align}

Alternatively, one writes 
\begin{align}\label{ftheta_boundedsmooth_}
    f(\theta) &=
    \begin{cases}
        \theta_1 \quad\text{if }  \theta > 2 \\ 
        \frac{1}{4} \left(\frac{4}{3}+ 2 (\theta_1+\theta_2) +(\theta_1-\theta_2)^2 -\frac{1}{6}(\theta_1-\theta_2)^3 \right) \quad\text{if } 2 \geq \theta \geq 0 \\ 
        \frac{1}{4} \left(\frac{4}{3} +2(\theta_1 + \theta_2)+ (\theta_1-\theta_2)^2 + \frac{1}{6}(\theta_1-\theta_2)^3\right)  \quad\text{if } 0 \geq \theta \geq -2 \\
        \theta_2 \quad\text{if }  \theta < -2
    \end{cases}
\end{align}

The derivative of $f(\theta)$ (Equation \ref{ftheta_boundedsmooth}), $\frac{\partial}{\partial \theta}f(\theta) = (\frac{\partial}{\partial \theta_1}f(\theta), \frac{\partial}{\partial \theta_2}f(\theta))$ corresponds to the probabilities of the arg max, $(P_\gamma \left(\arg \max _i \{ \theta_i+ \gamma_i \} = 1 \right),P_\gamma \left(\arg \max _i \{ \theta_i+ \gamma_i \} = 2 \right))$ :

\begin{align}\label{maximizer_prob_boundedsmooth}
    \frac{\partial}{\partial \theta}f(\theta) &=
    \begin{cases}
        (1,0) \quad\text{if }  -\theta < -2 \\ 
        (\frac{1}{2}+ \frac{1}{2}\theta+ \frac{1}{8}\theta^2, \frac{1}{2}- \frac{1}{2}\theta - \frac{1}{8}\theta^2) \quad\text{if } 0 \leq -\theta \leq 2 \\
       (\frac{1}{2}+\frac{1}{2}\theta-\frac{1}{8}\theta^2,\frac{1}{2}-\frac{1}{2}\theta + \frac{1}{8}\theta^2 ) \quad\text{if } -2 \leq -\theta \leq 0 \\ 
       (0,1)\quad\text{if }  -\theta > 2
    \end{cases}
\end{align}

\begin{align}
    \frac{\partial}{\partial \theta}f(\theta) &=
    \begin{cases}
        (1,0) \quad\text{if }  \theta > 2 \\ 
        (\frac{1}{2}+\frac{1}{2}\theta-\frac{1}{8}\theta^2,\frac{1}{2}-\frac{1}{2}\theta + \frac{1}{8}\theta^2 ) \quad\text{if } 2 \geq \theta \geq 0 \\ 
        (\frac{1}{2}+ \frac{1}{2}\theta+ \frac{1}{8}\theta^2, \frac{1}{2}- \frac{1}{2}\theta - \frac{1}{8}\theta^2) \quad\text{if } 0 \geq \theta \geq -2 \\
       (0,1)\quad\text{if }  \theta < -2
    \end{cases}
\end{align}

Then, the partial derivatives $\frac{\partial}{\partial \theta}f(\theta) \in [0,1]$ and sum to $1$, as expected.
\begin{equation}
    \frac{\partial}{\partial \theta_1}f(\theta)+ \frac{\partial}{\partial \theta_2}f(\theta) = 1
\end{equation} 

\begin{enumerate}
\item \textbf{Case: $0 \leq -\theta \leq 2$} 
\begin{align}
    \frac{\partial}{\partial \theta}f(\theta) &=(\frac{1}{4}(2 + 2\theta_1-2\theta_2 +\frac{3\theta_1^2-6\theta_1\theta_2+3\theta_2^2}{6}), \frac{1}{4}(2 + 2\theta_2-2\theta_1 + \frac{-3\theta_1^2+6\theta_1\theta_2-3\theta_2^2}{6})) \\ 
    %&= (\frac{1}{2}+ \frac{1}{2}(\theta_1-\theta_2)+ \frac{1}{8}(\theta_1-\theta_2)^2, \frac{1}{2}- \frac{1}{2}(\theta_1-\theta_2) - \frac{1}{8}(\theta_1-\theta_2)^2) \\
    &=(\frac{1}{2}+ \frac{1}{2}\theta+ \frac{1}{8}\theta^2, \frac{1}{2}- \frac{1}{2}\theta - \frac{1}{8}\theta^2)
\end{align}

The global minimum of the derivative $f(\theta)$ w.r.t. $\theta_1$,  $ \min \frac{\partial}{\partial \theta_1}f(\theta) = 0$ for $\theta = -2$, since
\begin{equation}
    \frac{\partial \frac{1}{4}(2 + 2\theta_1-2\theta_2 +\frac{3\theta_1^2-6\theta_1\theta_2+3\theta_2^2}{6})}{\partial \theta_1}= 
    \frac{1}{4}(2 +\theta_1 -\theta_2),
\end{equation}
in which case $\frac{\partial}{\partial \theta_2}f(\theta) = 1$.
The global maximum of the derivative $f(\theta)$ w.r.t. $\theta_1$,  $ \max \frac{\partial}{\partial \theta_1}f(\theta) = \frac{1}{2}$ for $-\theta = 0$, in which case $\frac{\partial}{\partial \theta_2}f(\theta) = \frac{1}{2}$.

The global maximum of the derivative $f(\theta)$ w.r.t. $\theta_2$,  $ \max \frac{\partial}{\partial \theta_2}f(\theta)) = 1$ for $\theta_2 = \theta_1+2$, since
\begin{equation}
    \frac{\partial \frac{1}{4}(2 + 2\theta_2-2\theta_1 + \frac{-3\theta_1^2+6\theta_1\theta_2-3\theta_2^2}{6}))}{\partial \theta_2}= 
    \frac{1}{4}(2 +\theta_1 -\theta_2),
\end{equation}
in which case $\frac{\partial}{\partial \theta_1}f(\theta) = 0$.
The global minimum of the derivative $f(\theta)$ w.r.t. $\theta_2$,  $ \max \frac{\partial}{\partial \theta_2}f(\theta) = \frac{1}{2}$ for $-\theta = 0$, in which case $\frac{\partial}{\partial \theta_1}f(\theta) = \frac{1}{2}$.
\end{enumerate}

\section{Experiments}
We use a $16$GB 6-Core Intel Core i7 CPU in both experiments.   

\subsection{Approximating discrete distributions}\label{app_exp_apprpx}
\begin{figure}[th]
\centering
\begin{subfigure}[t]{.24\linewidth}
\includegraphics[width=\linewidth]{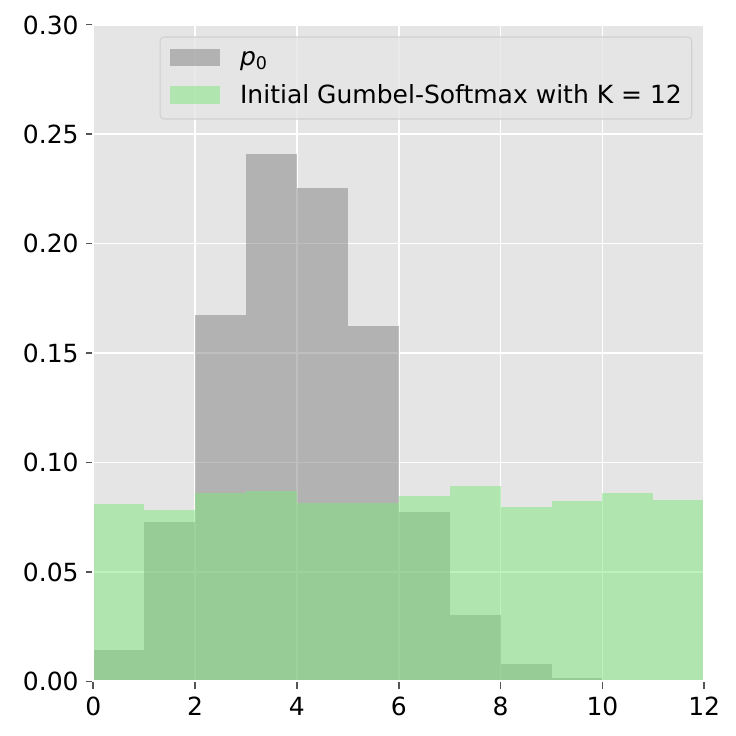}
\end{subfigure}
\begin{subfigure}[t]{.24\linewidth}
\includegraphics[width=\linewidth]{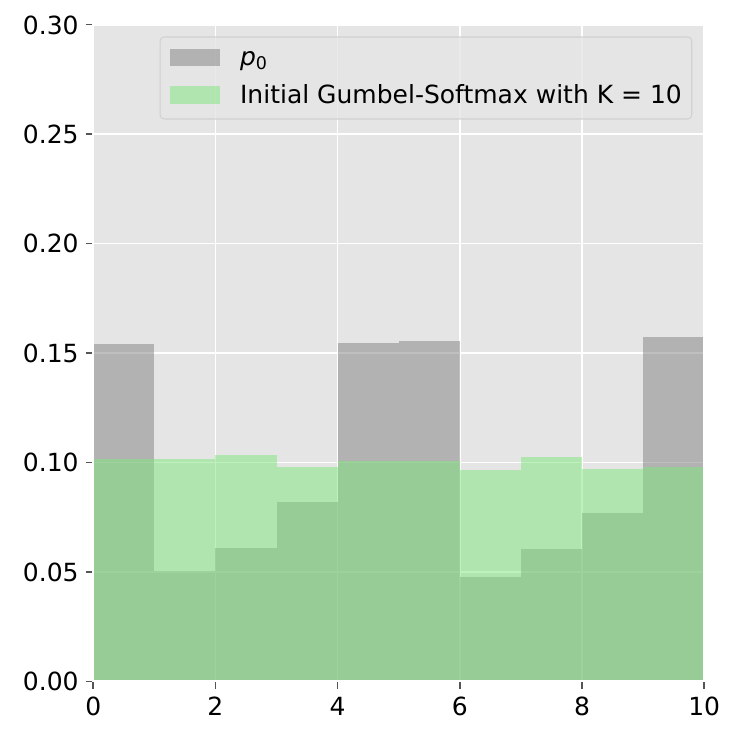}
\end{subfigure}
\begin{subfigure}[t]{.24\linewidth}
\includegraphics[width=\linewidth]{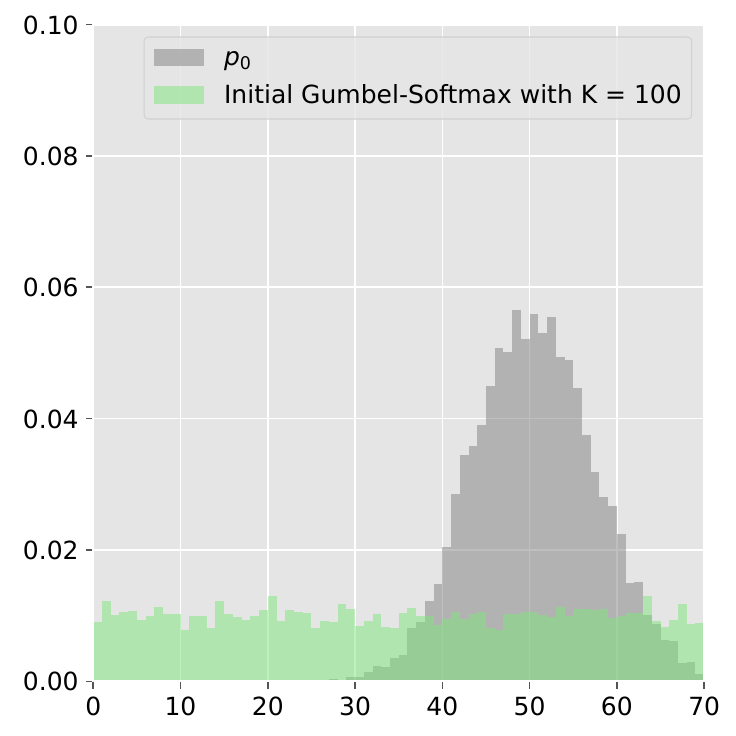}
\end{subfigure}
\begin{subfigure}[t]{.24\linewidth}
\includegraphics[width=\linewidth]{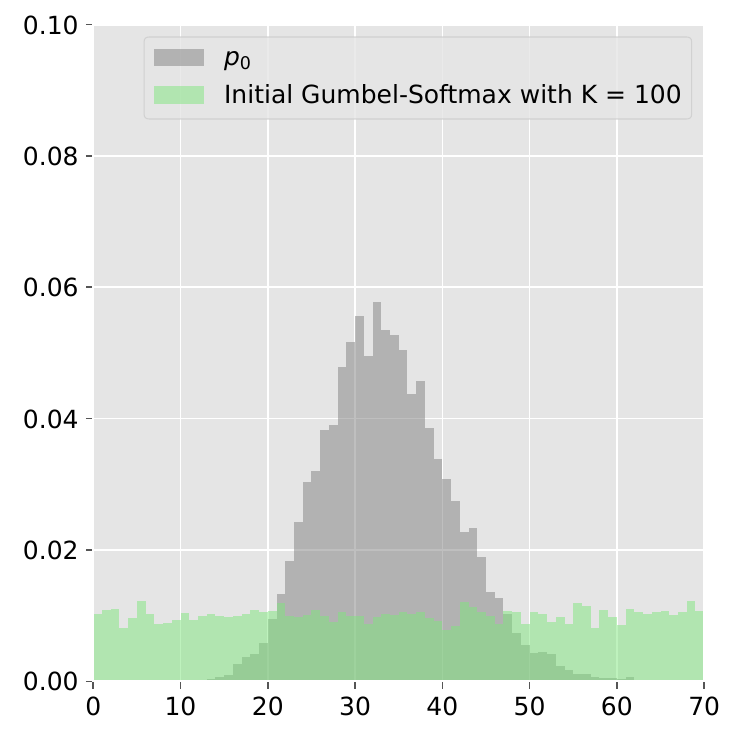}
\end{subfigure}

\begin{subfigure}[t]{.24\linewidth}
\includegraphics[width=\linewidth]{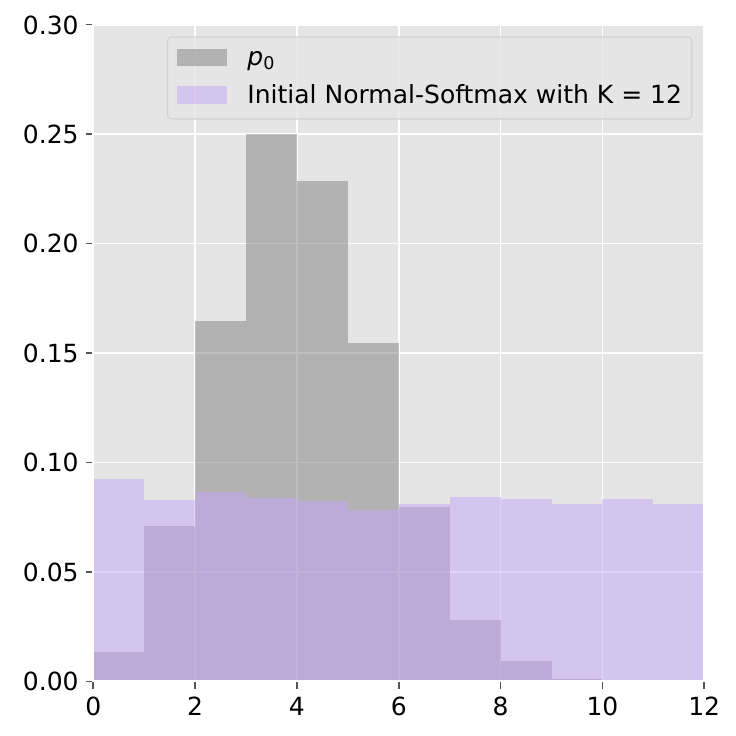}
\caption{a binomial distribution}
\label{fig:binomial_init}
\end{subfigure}
\begin{subfigure}[t]{.24\linewidth}
\includegraphics[width=\linewidth]{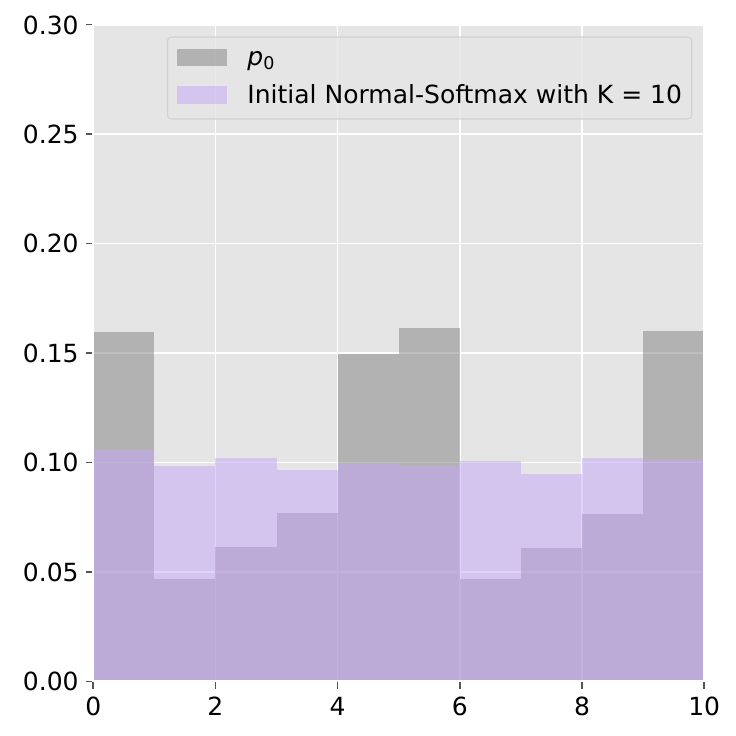}
\caption{a discrete distribution}
\label{fig:discrete_init}
\end{subfigure}
\begin{subfigure}[t]{.24\linewidth}
\includegraphics[width=\linewidth]{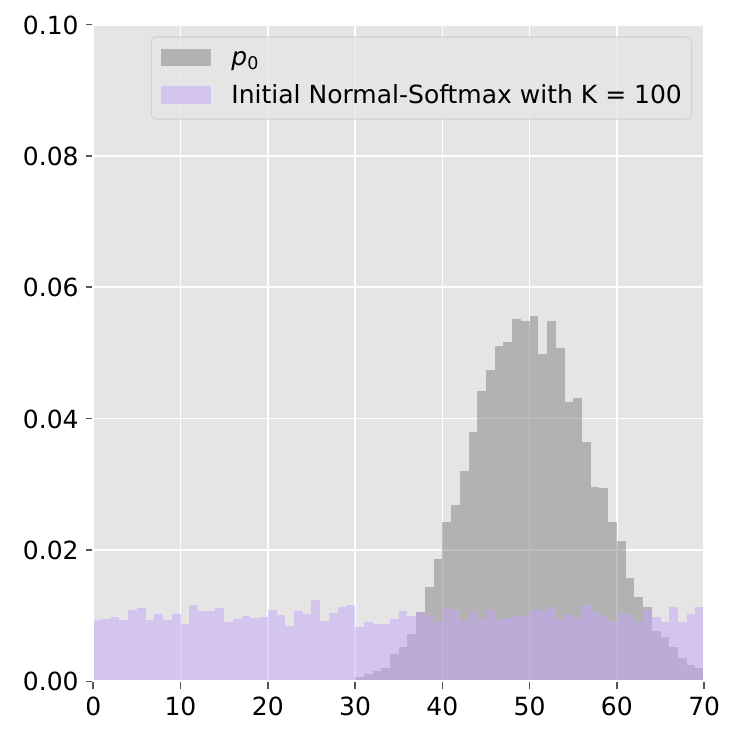}
\caption{a Poisson distribution}
\label{fig:Poisson_init}
\end{subfigure}
\begin{subfigure}[t]{.24\linewidth}
\includegraphics[width=\linewidth]{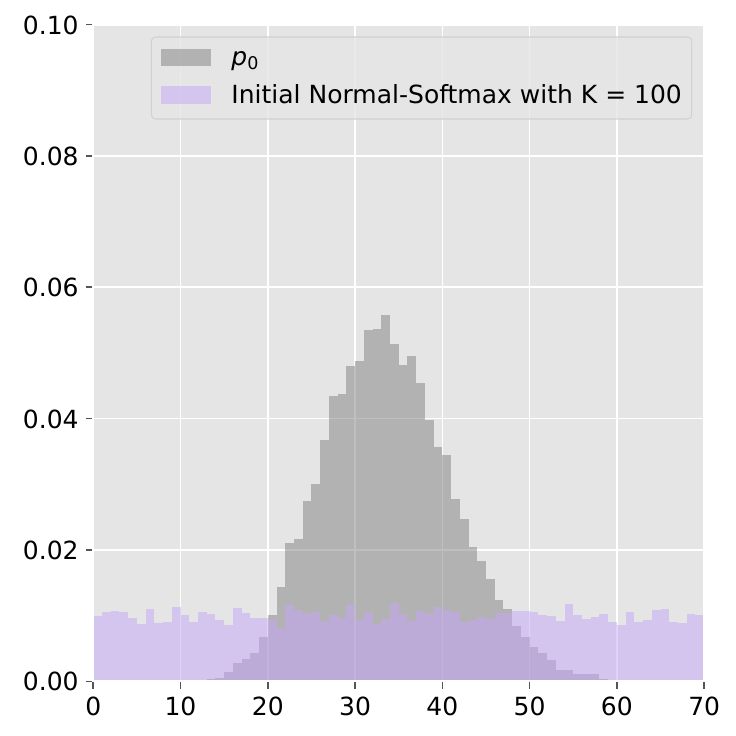}
\caption{a negative binomial distribution}
\label{fig:negativebinomial_init}
\end{subfigure}
\caption{Visualization of the uniform initialization of the fitted Gumbel-Softmax (top row) and Normal-Softmax (bottom row) for the target discrete distributions: \ref{fig:binomial_init} a target binomial distribution $p_0$ with $n = 12, p = 0.3$, \ref{fig:discrete_init} a discrete distribution with $p = (\frac{10}{68},\frac{3}{68}, \frac{4}{68}, \frac{5}{68}, \frac{10}{68}, \frac{10}{68}, \frac{3}{68}, \frac{4}{68}, \frac{5}{68}, \frac{10}{68})$, \ref{fig:Poisson_init} a Poisson distribution with $\lambda=50$, and \ref{fig:negativebinomial_init} a negative binomial distribution with $r=50, p=0.6$.}
\label{fig:approx_dist_fitting_init}
\end{figure}

\begin{figure}[h]
\centering
\begin{subfigure}[c]{\textwidth}
  \centering
  \includegraphics[width=0.3\linewidth]{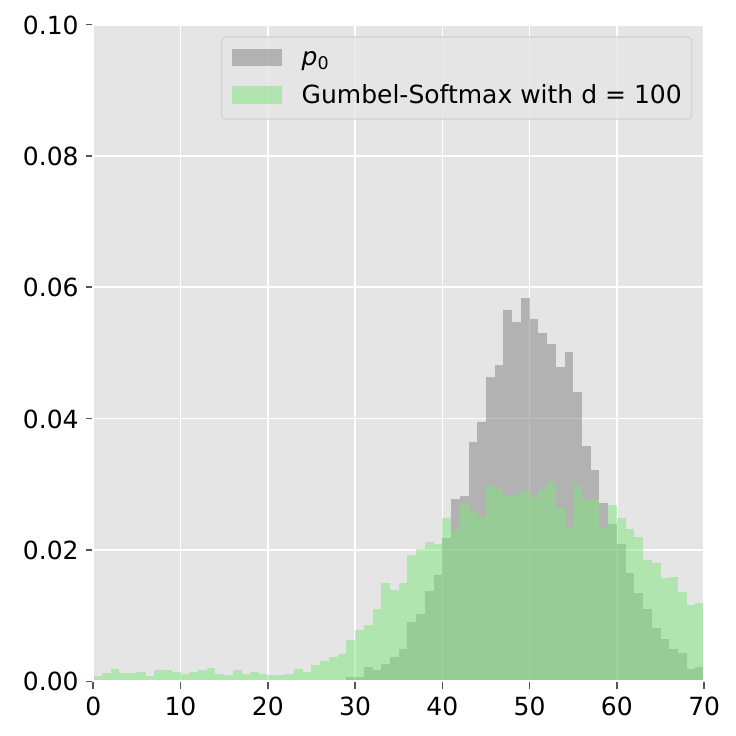}
  \includegraphics[width=0.3\linewidth]{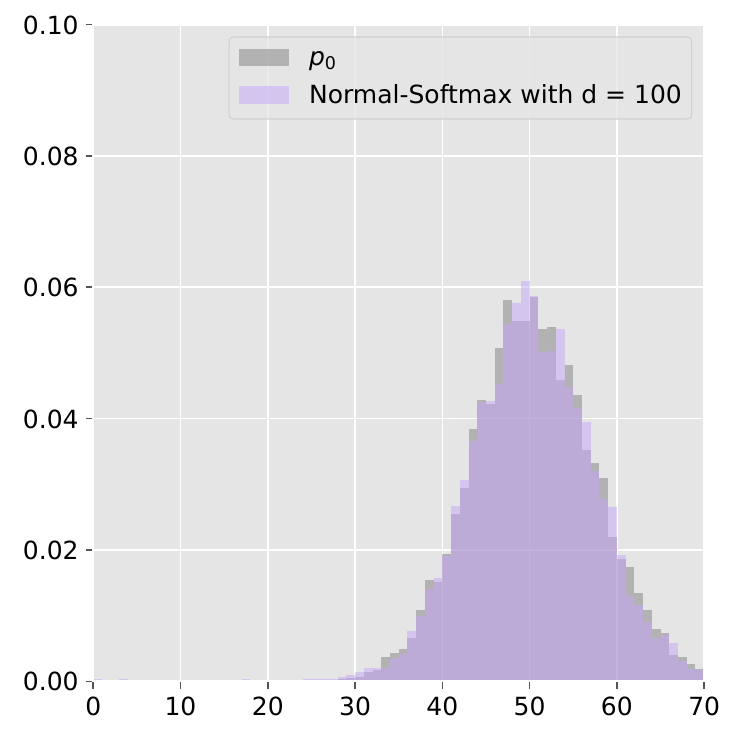}
  \includegraphics[width=0.32\linewidth]{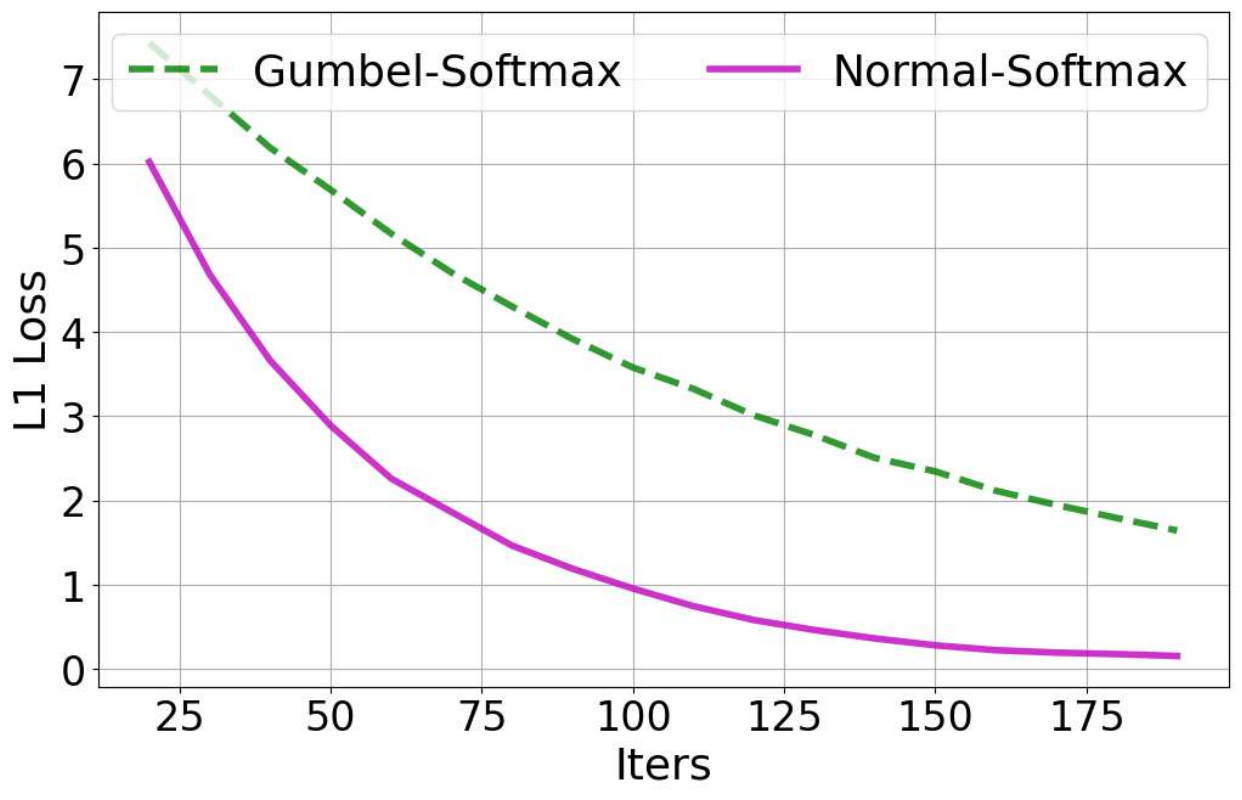}
  \caption{Approximation of a Poisson distribution.}
\label{fig:Case3_discreteinfinitesupport}
\end{subfigure}
\vskip\baselineskip
\begin{subfigure}[c]{\textwidth}
  \centering
  \includegraphics[width=0.3\linewidth]{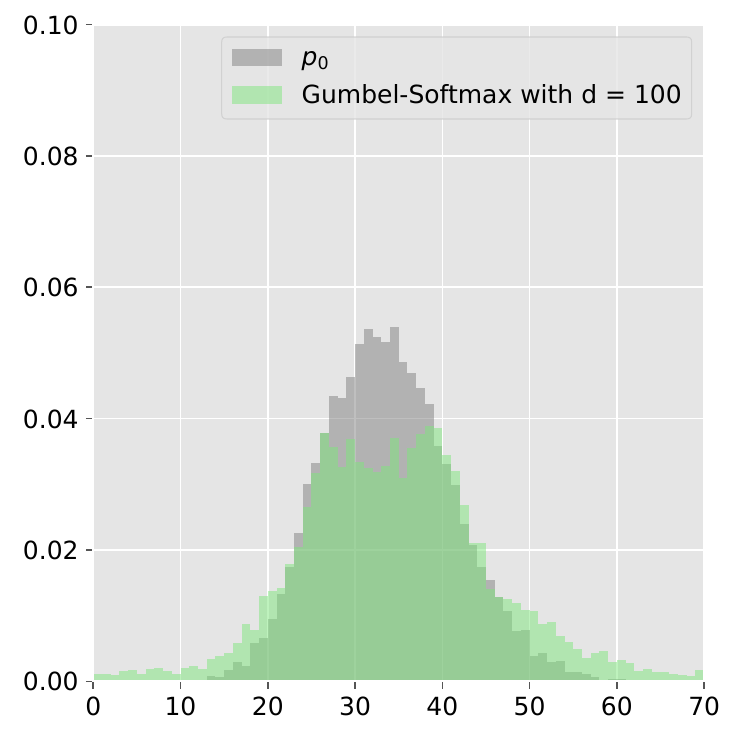}
  \includegraphics[width=0.3\linewidth]{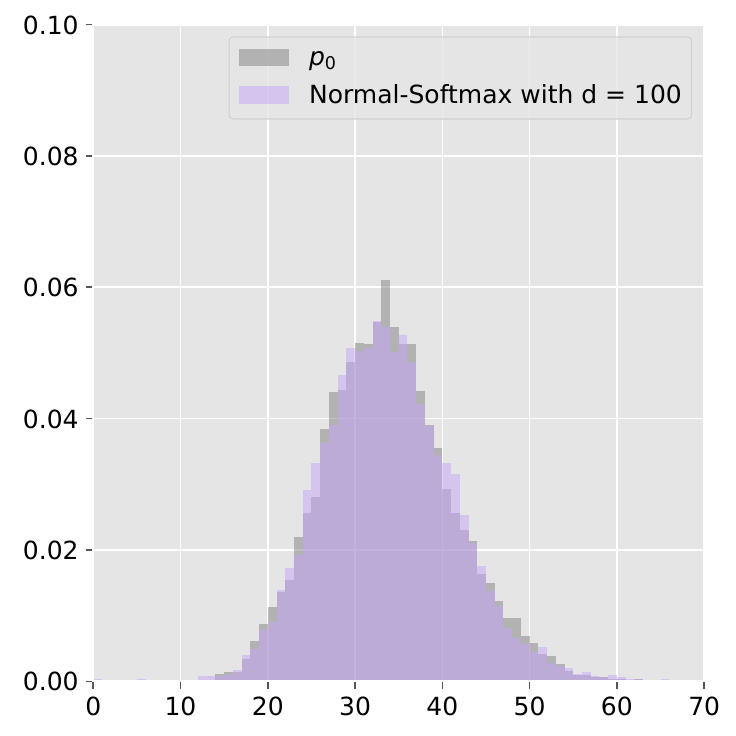}
  \includegraphics[width=0.32\linewidth]{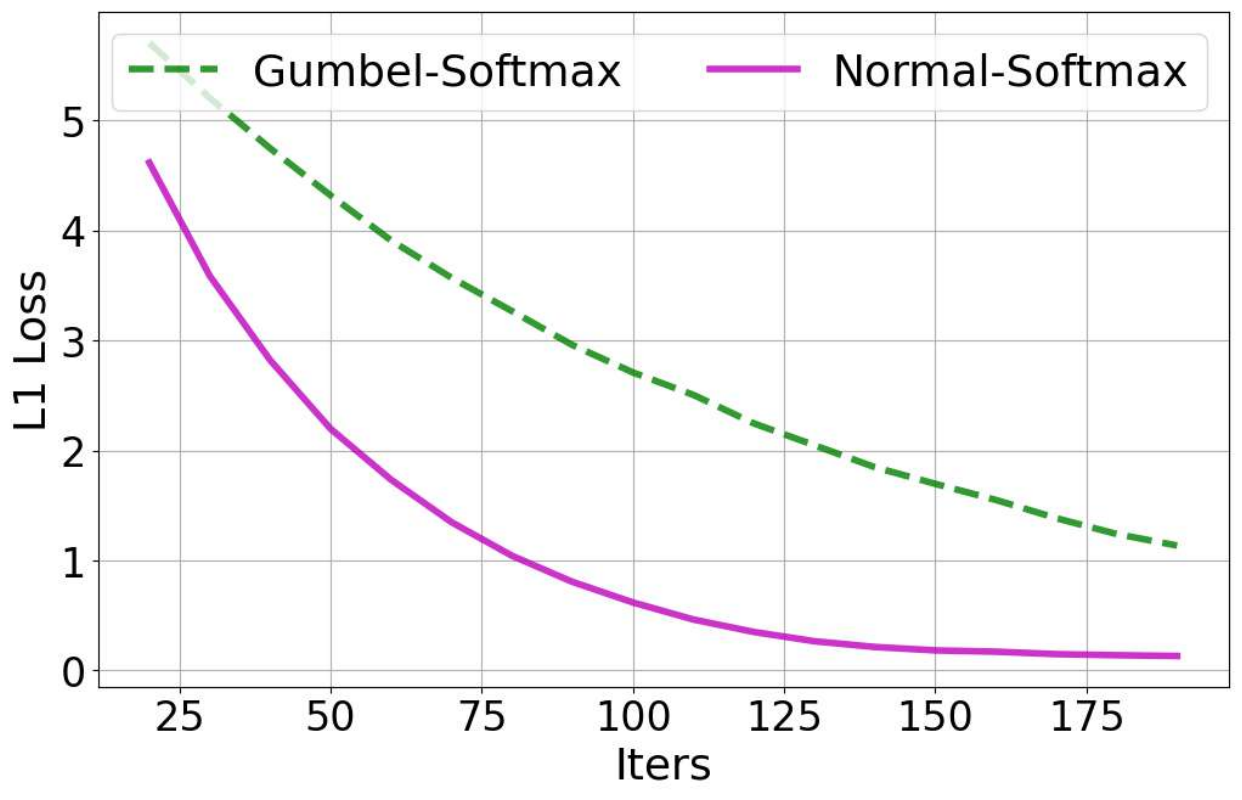}
  \caption{Approximation of a negative binomial distribution.}
\label{fig:Case4_discreteinfinitesupport}
\end{subfigure}
\caption{Gumbel-Softmax and Normal-Softmax approximation with dimension $d=100$ of target discrete distributions $p_0$ with countably infinite support. The $L1$ objective over learning iterations is depicted on the right.}
\label{fig:approx_dist_infinitesupport}
\end{figure}

Our experiments are based on the publicly available code of  \citet{NEURIPS2020_90c34175}. 
In all experiments, a thousand samples from a target distribution $p_0$ are sampled to approximate its probability density function parameters, based on the $L_1$ objective function. Optimization is based on the Adam optimizer \citep{kingma2017adam} with a learning rate $1.e-2$. The fitted parameters are initialized uniformly, as visualized in Figure \ref{fig:approx_dist_fitting_init}.
Figure \ref{fig:approx_dist_infinitesupport} shows results for fitting discrete distributions with countably
infinite support: a Poisson distribution with $\lambda=50$,  and a negative binomial distribution with $r=50, p=0.6$ respectively. The Normal-Softmax distribution better approximates both distributions and exhibits faster
convergence than the Gaussian-Softmax.

Table \ref{table:aproxx-dist-l1} shows the average and standard deviation $L_1$ objective between the approximated probability density function and the four target discrete distributions of both models after $300$ iterations, computed over the dimension $d$ of the fitted models. The approximation based on the Normal-Softmax probability model achieves lower mean and standard deviation $L_1$ in all cases.

\begin{table}[h]
\caption{$L_1$ average and standard deviation between the target and approximated probability density function of various target discrete distributions of the Normal-Softmax and Gumbel-Softmax probability models.}
\label{table:aproxx-dist-l1}
\begin{center}
\begin{tabular}{lll}
\toprule
Target Distribution & \textbf{Normal-Softmax} & \textbf{Gumbel-Softmax} \\ \hline
Discrete & 0.026 $\pm$ 0.002 & 0.027$\pm$0.002 \\ \hline
Binomial & 0.036$\pm$0.003 & 0.177 $\pm$0.014 \\ \hline
Poisson & 0.090$\pm$0.001 & 0.618 $\pm$0.007 \\ \hline
Negative Binomial & 0.083 $\pm$ 0.001 & 0.417 $\pm$ 0.004 \\ \hline
\bottomrule
\end{tabular}
\end{center}
\end{table}

\subsection{Variational inference}
This experiment is based on the publicly available implementation of the Gumbel-Softmax-based implementation of the discrete VAE in \href{https://github.com/GuyLor/Direct-VAE}{Direct-VAE}. Optimization is based on the Adam optimizer \citep{kingma2017adam} with a learning rate of $1.e-3$. Batch size is set to  $100$. 
We use the regular train/ test splits and follow previous research splits (e.g., as in \citet{lorberbom2019direct}). The MNIST and the Fashion-MNIST datasets' training set comprises $60,000$ images, and the test set comprises $10,000$ images. For the Omniglot, the training set comprises $24,345$ images, and the test set comprises $8,070$ images.

\begin{figure}[th]
\centering
\begin{subfigure}[b]{.32\textwidth}
\includegraphics[width=\linewidth]{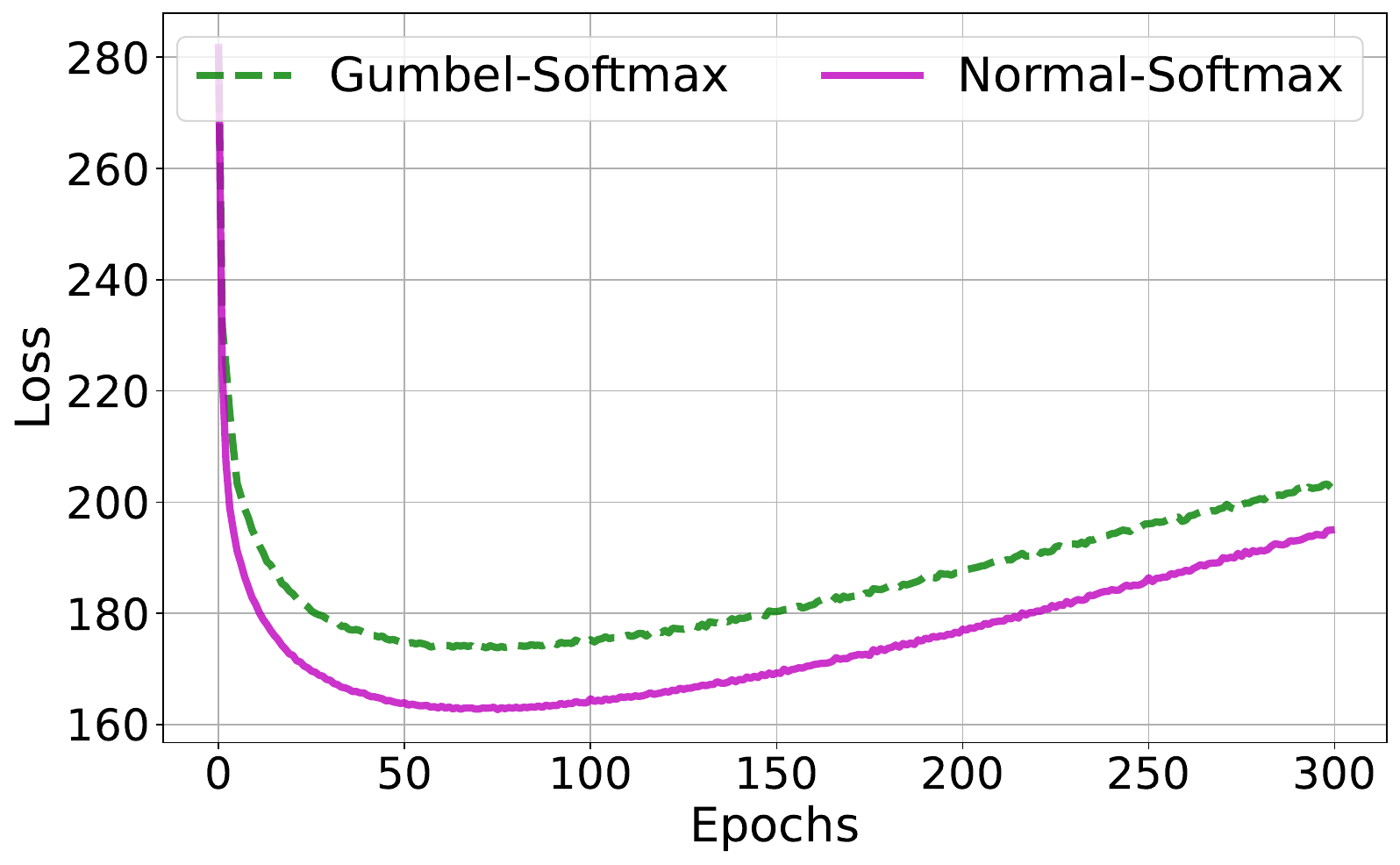}
\end{subfigure}
\begin{subfigure}[b]{.32\textwidth}
\includegraphics[width=\linewidth]{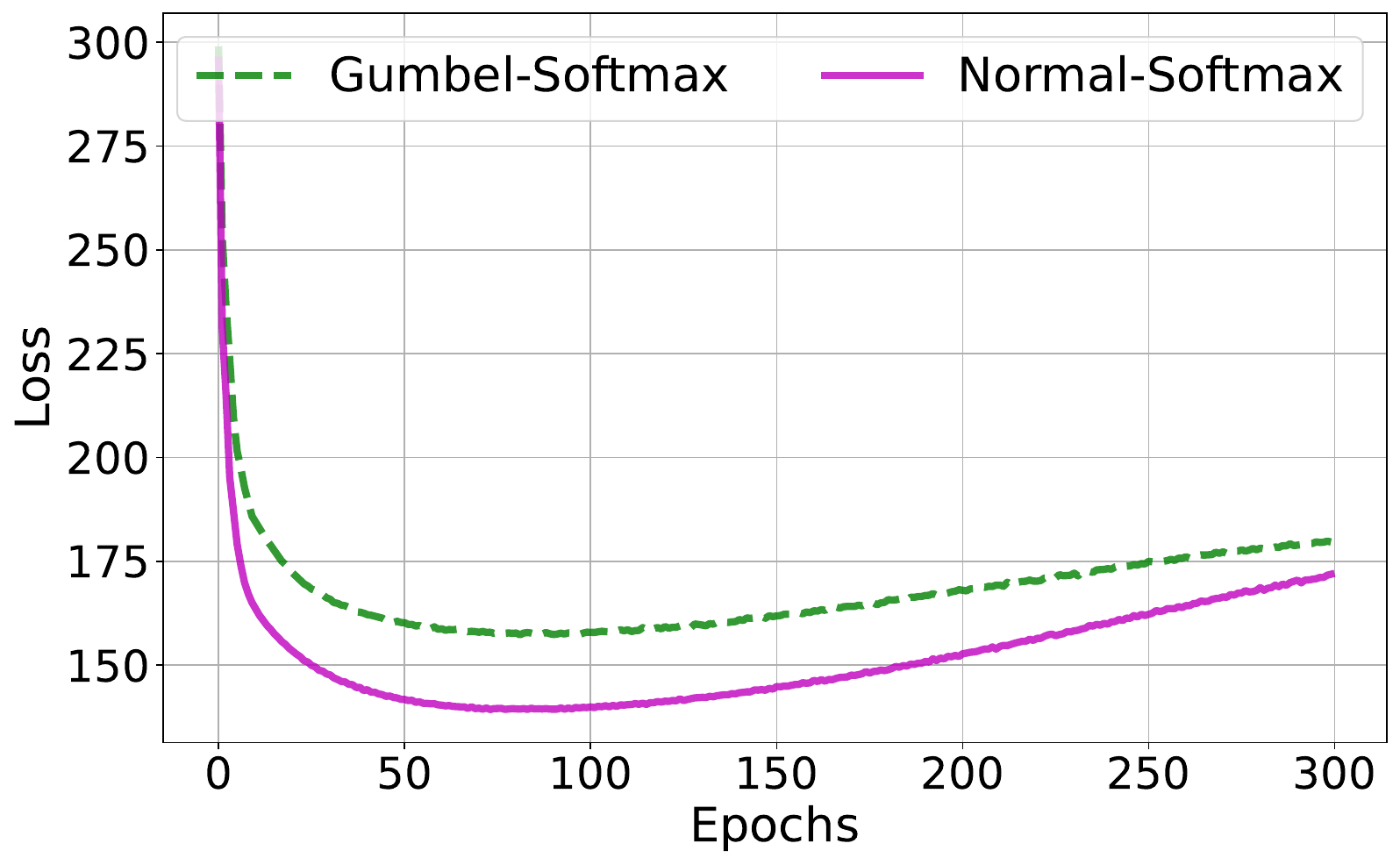}
\end{subfigure}
\begin{subfigure}[b]{.32\textwidth}
\includegraphics[width=\linewidth]{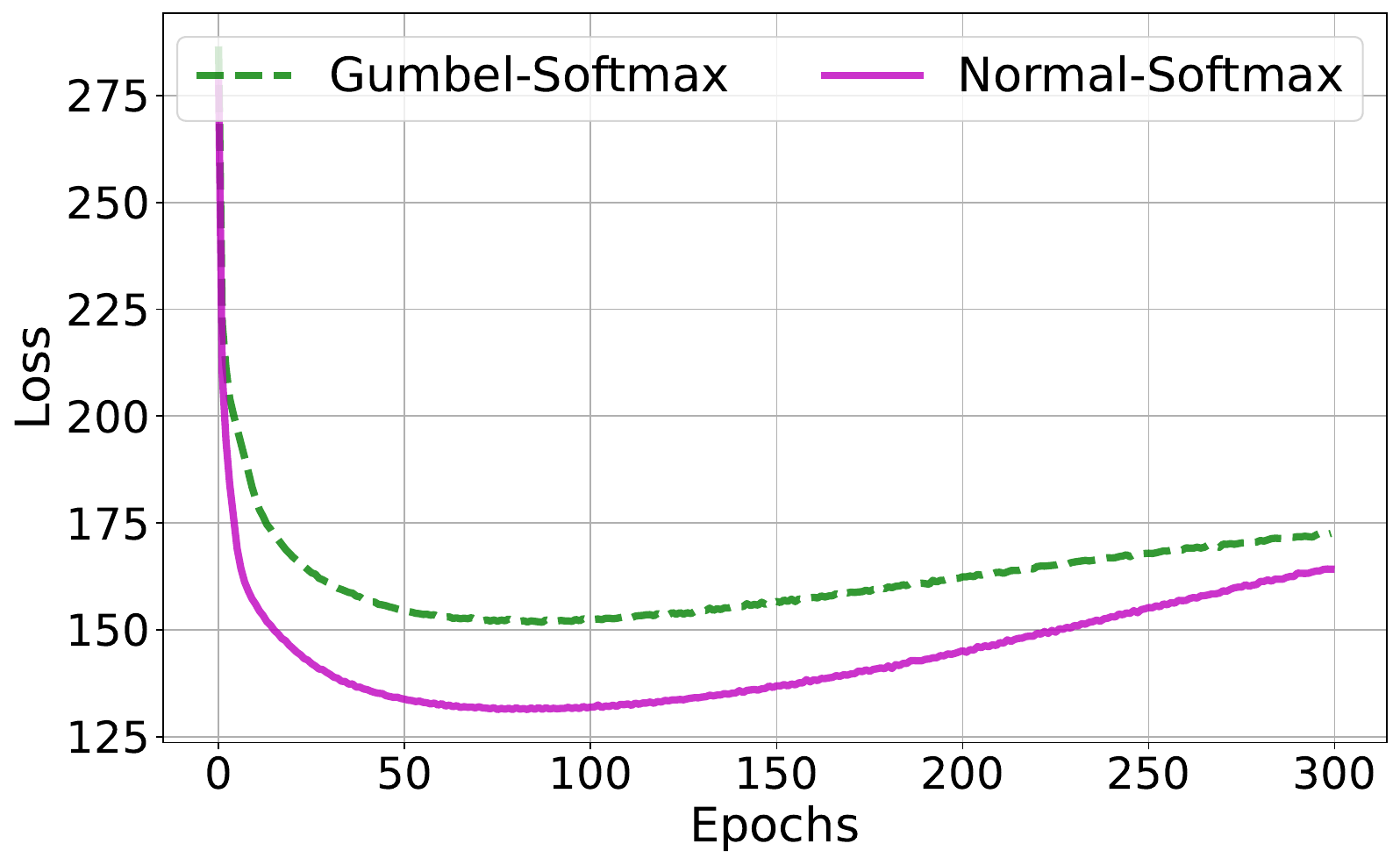}
\end{subfigure}
\caption{Categorical VAE with Perturb-Softmax training loss on the 
the Fashion-MNIST dataset with a $K$-dimensional categorical variable, $K\in [10,30,50]$.}
\label{Fashion-MNIST-perturb-sm}
\end{figure}

Figure \ref{Fashion-MNIST-perturb-sm} shows the results of the variational inference experiment for the Fashion-MNIST dataset \citep{xiao2017fashionmnist} with $N=10$ discrete variables, each is a $K$-dimensional categorical variable, $K\in [10,30,50]$. 

%\newpage
%\input{sections/Checklist}

\end{document}